\documentclass[journal]{IEEEtran}
\ifCLASSINFOpdf
\else
\fi
\usepackage{times}
\usepackage{graphicx}
\usepackage{amssymb}
\usepackage{clrscode}
\usepackage{algorithm}
\usepackage{algorithmic}
\usepackage{amstext}
\usepackage{multirow}
\usepackage{array}

\usepackage{cite}
\usepackage{amsmath}
\usepackage{algorithmic}
\usepackage{bm}
\usepackage{balance}

\usepackage{multirow}

\usepackage{subfigure}

\usepackage{xspace}
\makeatletter
\DeclareRobustCommand\onedot{\futurelet\@let@token\@onedot}
\def\@onedot{\ifx\@let@token.\else.\null\fi\xspace}

\def\eg{\emph{e.g}\onedot} 
\def\ie{\emph{i.e}\onedot} 
 
\def\etc{\emph{etc}\onedot} 
 
\def\etal{\emph{et al}\onedot}
\makeatother

\begin{document}
%
\title{A Large-scale Varying-view RGB-D Action Dataset for Arbitrary-view Human Action Recognition}
%
%
%

\author{Yanli Ji, 
        Feixiang Xu, Yang Yang, Fumin Shen, Heng Tao Shen,
        and~ Wei-Shi Zheng
\thanks{Yanli Ji, Feixiang Xu, Yang Yang, Fumin Shen, and~Heng Tao Shen are with Center for Future Media, School of Computer Science and Engineering, University of Electronic Science and Technology of China, Chengdu, China. e-mail: yanliji@uestc.edu.cn.}
\thanks{Wei-Shi Zheng is with School of Data and Computer Science,Sun Yat-sen University, China.}
\thanks{Corresponding author: Heng Tao Shen.}
\thanks{Manuscript received XX, 2018; revised XX, 2018.}}

%
%

\markboth{~Vol.~, No.~, April~2019}%
{Shell \MakeLowercase{\textit{et al.}}: Bare Demo of IEEEtran.cls for IEEE Journals}
%



\maketitle

\begin{abstract}
Current researches of action recognition mainly focus on single-view and multi-view recognition, which can hardly satisfies the requirements of human-robot interaction (HRI) applications to recognize actions from arbitrary views. The lack of datasets also sets up barriers. To provide data for arbitrary-view action recognition, we newly collect a large-scale RGB-D action dataset for arbitrary-view action analysis, including RGB videos, depth and skeleton sequences. The dataset includes action samples captured in 8 fixed viewpoints and varying-view sequences which covers the entire $360^\circ$ view angles. In total, 118 persons are invited to act 40 action categories, and 25,600 video samples are collected. Our dataset involves more participants, more viewpoints and a large number of samples. More importantly, it is the first dataset containing the entire $360^\circ$ varying-view sequences. The dataset provides sufficient data for multi-view, cross-view and arbitrary-view action analysis. Besides, we propose a View-guided Skeleton CNN (VS-CNN) to tackle the problem of arbitrary-view action recognition. Experiment results show that the VS-CNN achieves superior performance.
\end{abstract}


%
\IEEEpeerreviewmaketitle

\section{Introduction}

\begin{figure}
\begin{center}
\includegraphics[width=0.9\linewidth]{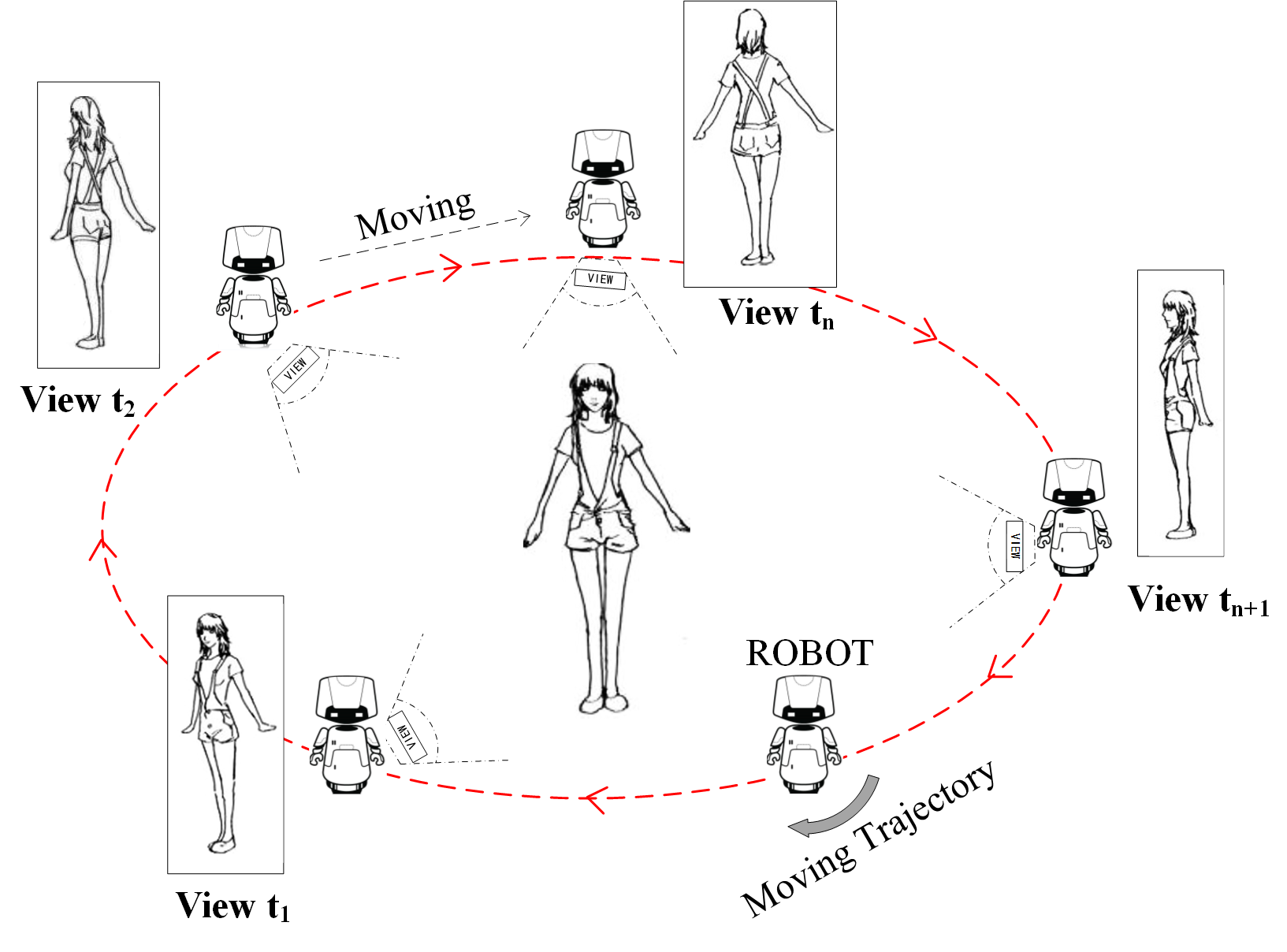}
\end{center}
   \caption{Arbitrary-view actions in the HRI application. Robots move around the human and are expected to understand human behaviors in arbitrary views. }
\label{fig:MovingViewHRI}
\end{figure}

Human action recognition is widely applied in public surveillance, image/video captioning and human-robot interaction~\cite{TIP2017Event,CVPRJirongrong,yi2018describing}, \etc. Approaches for action recognition has developed from silhouettes~\cite{ZhuLSRF2013,GaoCollab2017}, local features~\cite{TIP2016Semantic,TIP2014image,wu2018sequence} to depth features~\cite{ZhengSoft2016,CookingActivity2012}, and skeletons~\cite{LiWWHL2017,Ji2017One}. Existing researches focus on single-view (mostly in the front viewpoint) and multi-view action recognitions~\cite{hash2017}. However, they can hardly satisfy the demand of robots to recognize human actions in arbitrary views for Human-robot interaction (HRI) applications. Taking a service robot at home (shown in Fig.~\ref{fig:MovingViewHRI}) as an example, it freely moves to anywhere and interacts with family members. During moving, the robot captures human actions in any viewpoints, and it is certainly expected to understand human behaviors in arbitrary viewpoints. However, the arbitrary-view human action recognition is still a big challenging problem. On the one hand, view changes lead to action occlusions and pose variances. On the other hand, there are few datasets for arbitrary-view action recognition.

There have been datasets developed for multi-view action recognition~\cite{IXMAS2006}. RGB information is used for the multi-view action recognition~\cite{IXMAS2007,LiuTransfer2011}. With the development of depth sensors, datasets containing RGB-D information were presented, such as $Act4^2$, Multiview 3D Event, Northwestern-UCLA, and UWA3D Multiview, and the NTU RGB+D action dataset~\cite{ChengACT42012,Wei4DHOI2013,WangMSTAOG2014,RahmaniUWA3D2016,ShahroudyNTU2016}. For the arbitrary-view recognition, it is expected to use action samples captured in wide-range views for model training. However, almost all existing datasets were captured in limited viewpoints. Taking advantage of the mocap motion in the CMU action dataset\footnote{http://mocap.cs.cmu.edu}, training datasets including action samples of various viewpoints were generated and used to train classifiers for multi-view action recognition~\cite{Gupta3Dpose2014,RahmaniNKTM2015,RahmaniNovelV2016}. Nonetheless, the dataset generation suffers from expensive computational cost, and the mocap motion dataset is required to cover a large number of action categories, which is also a difficult problem.
To solve the problem of lacking suitable data, we present a large-scale RGB-D action dataset which contains varying-view sequences covering the entire $360^\circ$ view angels. The dataset provides sufficient samples for the arbitrary-view action recognition.


Many research endeavors have been dedicated to solving the problem of multi-view action recognition. Transfer learning methods were adopted to transfer the knowledge from one viewpoint to other viewpoints~\cite{Rybok2011,Li2012,Zhang2013,M2Ibenchmark2016,GaoCollab2017}, or to transfer feature knowledge from the benchmark dataset to test datasets~\cite{LiuTransfer2011}. Since action sequences observed in varying viewpoints easily suffer occlusions, temporal motion is used for view-invariance action representation~\cite{ShahroudyNTU2016,InteractionJi2015,Wei4DHOI2013}. A further solution is to learn spatial relationships of joints from 3D poses to construct view-invariant representations~\cite{RahmaniHOPC2014,WangActionlet2013,Ji2017One}. However, most of the existing approaches can only deal with small-range view changes. For action recognition with view changes, one solution is to seek a common representation for actions in different views. Liu~\etal~\cite{EnhancedSK2017} visualized a skeleton sequence into a color image for action representation and presented the SK-CNN approach to recognize actions, which is potent to weaken the difference between different views. Regarding our dataset containing full-circle views, current solutions cannot handle the recognition task. To cover the full-circle view, we separate the full-circle view into four view groups, and propose a View-guided Skeleton-CNN (VS-CNN) approach to recognize actions with large view changes.


In this paper, we newly collect a large-scale RGB-D action dataset for arbitrary-view action recognition. The dataset contains samples captured in 8 fixed viewpoints and varying-view sequences that cover the entire $360^\circ$ view angels. Samples captured in fixed viewpoints provide training data for the arbitrary-view recognition, and also may be used for the multi-view recognition. The dataset contains 40 fitness action categories, and 118 persons are invited to act these actions. In total 83 hours' RGB videos are collected, and depth image sequences, skeleton sequences have similar frame numbers with RGB videos.
Moreover, we propose a baseline, termed View-guided Skeleton-CNN (VS-CNN) to tackle these problems. The model consists of a view-group prediction module and four classifiers corresponding to four view groups. The view-group prediction module guides the training of classifiers through separating action samples to four view groups, and driving the training of corresponding classifiers. Finally, a weighted fusion is performed on the four classifiers, and the SoftMax classifier is used to classify fused features to corresponding action categories. Since view groups overlap each other, the VS-CNN learns a common representation of actions in different view groups. In summary, our major contributions include:

$\bullet$
We present a large-scale RGB-D action dataset for arbitrary-view action recognition, which includes 118 subjects and 8 fixed viewpoints. To the best of our knowledge, this is one of the first datasets covering the entire $360^\circ$ varying-view sequences.

$\bullet$
To tackle the arbitrary-view action recognition problems, we propose the VS-CNN, which overcomes the gap of action recognition in large view ranges.

$\bullet$
The proposed approach is extensively evaluated on our collected dataset, and the promising performance validates the efficacy of both the approach and dataset.


\section{Related work}
\label{sec:relatedWork}

\subsection{Multi-view action recognition with 2D features}
The same with general action recognition, the crucial problem of multi-view action recognition is also to learn an effective representation for actions. There had developed many local features for action representation in 2D videos and depth sequences~\cite{TPAMIfcvid,TIP2015Action,hu2017robust}, and they were introduced for multi-view action recognition~\cite{Yan4D2012,2DMulti2014,TIP2016FineG}. To learn effective features, Hu~\etal~\cite{JOULE2016} presented the JOULE model which explored the shared and feature-specific components from multiple feature channels,~\ie RGB, and skeleton features, as heterogeneous features for action recognition. Recent years, Convolution Neuron Networks (CNN) were introduced for 2D feature learning, and a series of effective networks were developed,~\ie ResNeXt~\cite{ResNeXtARX2018}. To include temporal information of action sequences, LRCN (Long-term Recurrent Convolutional Networks)~\cite{LRCNCVPR2015} were presented for action recognition.

Since view variance leading to human pose change in 2D videos and depth sequences, a series of approaches were proposed to solve the problem. Liu~\etal~\cite{LiuTransfer2011} used the bipartite graph partitioning to cluster vocabularies collected from two independent viewpoints by a bag of visual-words into visual-word clusters, which bridged the semantic gap of actions between different viewpoints. Moreover, Liu~\etal~\cite{M2Ibenchmark2016} built a transferable dictionary pair for feature transformation between the front view and side view actions, and a common representation was obtained in the two views. Though Local features are insensitive to viewpoint change in a small range, it suffers serious occlusions when a large view change occurs. Therefore, approaches that could be used to solve the problem of view change are required.


\subsection{Multi-view action recognition with 3D features}
The 3D information plays important roles in multi-view action recognition~\cite{InteractionJi2015,HOI2016}. Building bridges between a large collection dataset and test datasets, multi-view action recognition was realized by matching sequences of various viewpoints to data samples of the large collection dataset to reduce the gap between different viewpoints~\cite{Gupta3Dpose2014,RahmaniNKTM2015,RahmaniNovelV2016}.
However, a major limitation is the expensive computational cost for dataset generation, and the mocap motion dataset is also required to have numerous action categories.
A solution was to learn spatial relationships of 3D joints for a view-invariant action representation and recognition~\cite{ChengACT42012,RahmaniHOPC2014,WangActionlet2013}.
Moreover, Shahroudy~\etal~\cite{ShahroudyNTU2016} presented a Part-aware LSTM model (P-LSTM) which contained multiple parallel memory cells for body-part feature learning and one output gate for information sharing among body parts. The P-LSTM combined body-part context information, and provided a global representation for action recognition. Graph model was employed to model the 3D geometric relations for multi-view recognition~\cite{Wei4DHOI2013,WangMSTAOG2014}.
These high-level representations somewhat produced a common description in different viewpoints.

Moreover, some approaches transformed the 3D skeleton feature to 2D visual images, and took advantage of feature learning via CNN to achieve higher action recognition results. Kim~\etal~\cite{ResTCN2017} collected temporal skeleton trajectories and created frame-wise skeleton features concatenated temporally across the entire video sequence, and the Res-TCN was designed for action recognition. Liu~\etal~\cite{EnhancedSK2017} visualized skeleton motions of an action sequence to an enhanced color image, and a multi-stream CNN fusion model was used to recognize actions (SK-CNN). Yan~\etal~\cite{STGCN2018} modeled spatiotemporal skeletons using a Spatial-Temporal Graph Convolutional Networks (ST-GCN), which learned the importance of skeleton joints and assigned proper weights on graph convolution layers for action representation. Benefiting from effective feature extraction via CNN, these approaches had good performance with small-range view changes. In this paper, we propose a VS-CNN model to deal with action recognition in a large view ranges.

\section{Overview of Related Datasets}
\label{sec:OverviewDatabase}
Several multi-view human action datasets had been released. Weinland~\etal~\cite{IXMAS2006} released the IXMAS dataset containing RGB videos of human actions. The dataset was captured in five fixed viewpoints and contained 11 basic action categories, each performed by 10 actors. With the depth sensor Kinect V1, Cheng~\etal~\cite{ChengACT42012} presented the $ACT4^2$ action dataset including the RGB and depth information of 14 daily actions. 24 persons were invited to perform each action, and the dataset was captured in 4 fixed viewpoints. Wei~\etal~\cite{Wei4DHOI2013} built a multi-view 3D event dataset which included 8 event categories and 11 interacting object classes. RGB-D data of actions were captured using three stationary Kinect sensors. 8 persons were invited as participants in the data capture. Wang~\etal~\cite{WangMSTAOG2014} constructed the Northwestern-UCLA Multiview 3D event dataset which contained RGB, depth and skeleton data of 10 daily actions. Each action was performed by 10 participants, and data was captured in 3 fixed viewpoints. Rahmani~\etal~\cite{RahmaniHOPC2014} collected the UWA3D Multiview Activity Dataset in 4 viewpoints. The dataset contained 30 daily action categories, and each category was performed by 10 persons. Moreover, Shahroudy~\etal~\cite{ShahroudyNTU2016} presented a large-scale dataset, the NTU RGB+D action dataset. The dataset includes 60 daily actions, and totally 40 persons were invited for the data collection. Using 3 Kinect sensors, the dataset was captured in major 5 viewpoints. By changing camera-to-subject distances and camera heights, action data of 80 camera views were recorded.

Almost all existing datasets captured actions in limited viewpoints. It can hardly support the research of arbitrary-view action recognition for HRI applications. In addition, there are rare datasets including action samples captured in a very wide range of view angles and even continuously varying views.
To provide data for the arbitrary-view recognition, we simulate the HRI scenario and newly collect an action dataset which contains both action samples captured in fixed viewpoints and continuously varying-view action sequences. The varying-view sequence particularly covers the entire $360^\circ$ view angles, that is significantly different with existing datasets and beneficial for the evaluation of arbitrary-view action recognition.

\section{Varying-View RGB-D Action Dataset}
\label{sec:dataset}

\subsection{Database Description}
\begin{figure}
 \centering
 \subfigure[Capture setting A]{
 \label{fig:captureSetA} 
 \includegraphics[width=0.4\linewidth]{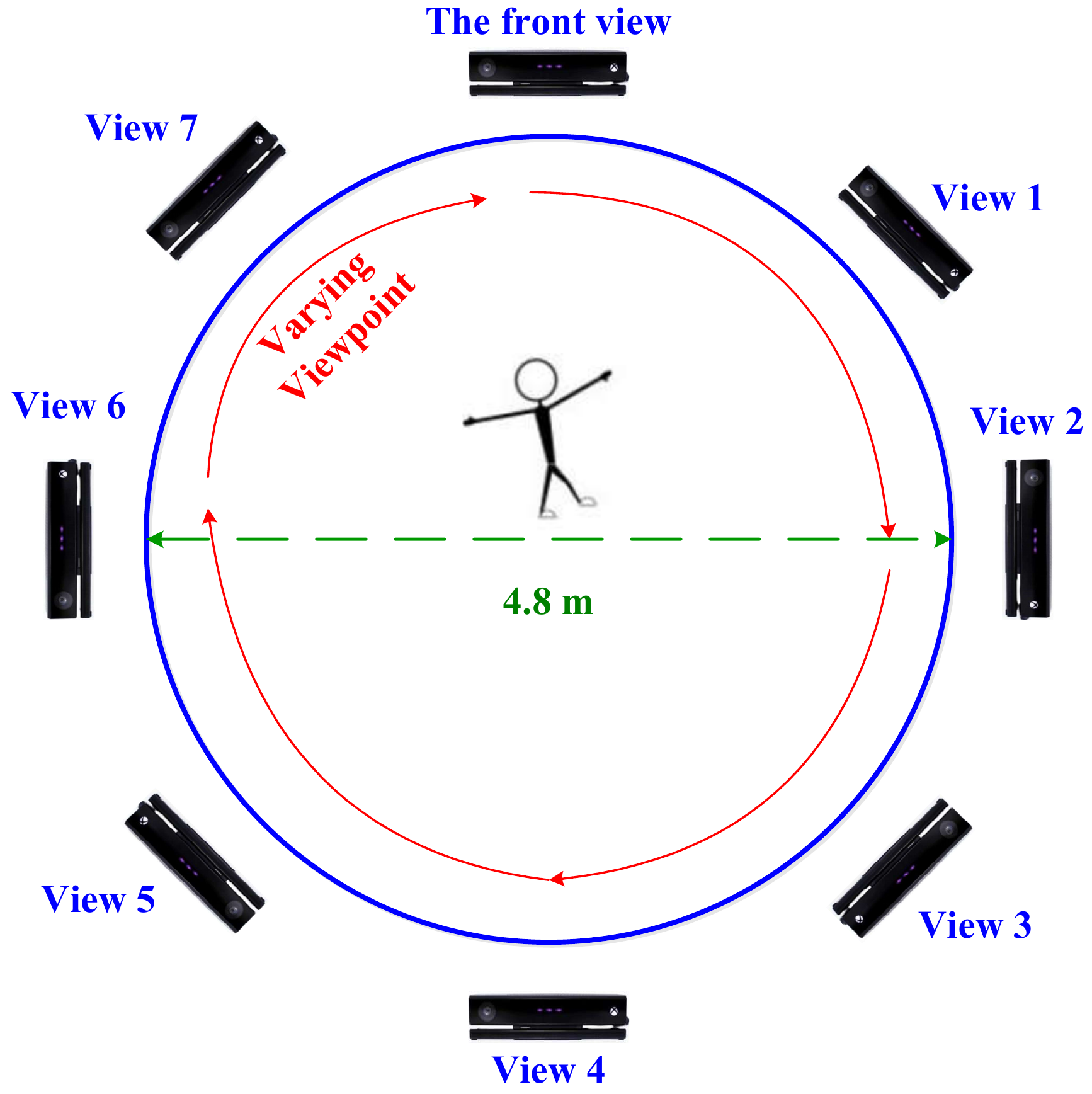}}
 \hspace{0.2in}
 \subfigure[Capture setting B and C]{
 \label{fig:captureSetB} 
 \includegraphics[width=0.4\linewidth]{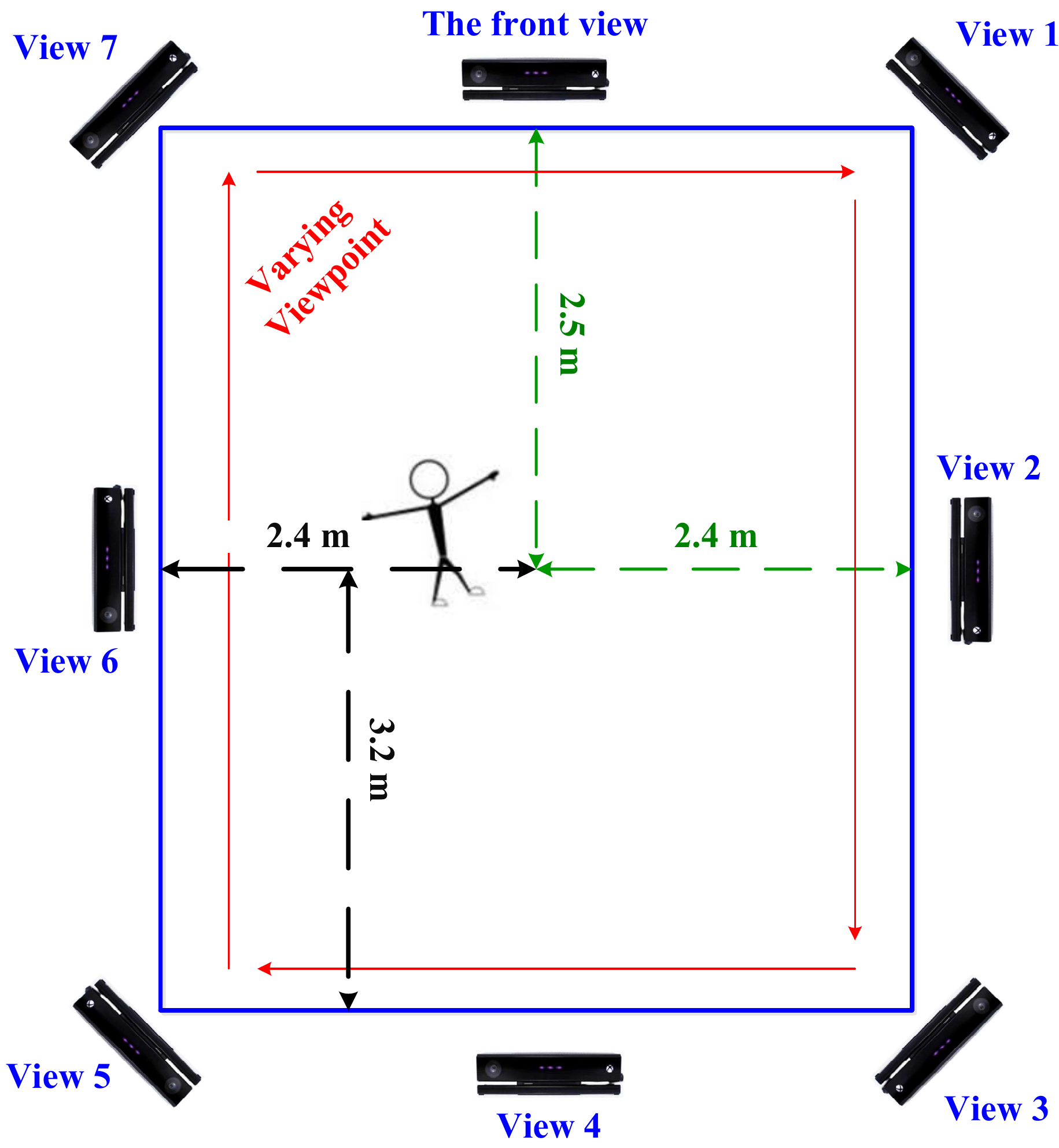}}
\caption{Three types of capture setting for our dataset construction. Sizes of setting B and C are shown with black dashes and green dashes in (b).}
 \label{fig:captureSet} 
 \end{figure}

The action dataset is collected using Microsoft Kinect v2 sensors. We use the sensor to capture 3 modality action data, \ie, RGB videos, Depth images, and skeleton sequences. For RGB videos, we record image frames with the resolution of $960\times540$ pixels. Depth images retain the maximum resolution of Kinect v2 sensors, $512\times424$ pixels, and 16-bit pixel values. Human skeletons contain 25 body joints per frame, and each joint is recorded as a 3D coordinate $(x,y,z)$ in the 3D space centered on the Kinect sensor. We show dataset capture settings in the Figure~\ref{fig:captureSet}. Camera positions indicate the 8 fixed viewpoints, and red arrows show the moving trajectory of sensors when we capture varying-view action sequences. During the data collection, subjects always face the Kinect sensor in the front viewpoint.~\footnote{The dataset has been released on  https://github.com/HRI-UESTC/CFM-HRI-RGB-D-action-database.} We collect fixed-viewpoint data to train classifiers for the arbitrary-view recognition because it is difficult to train a robust classifier using varying-view sequences due to fast varying views.

\begin{table*} \footnotesize
\begin{center}
\caption{Action categories in the Varying-View RGB-D Action dataset.}
\label{tab:category}
\begin{tabular}{|p{0.5cm}|p{2cm}|p{0.5cm}|p{2cm}|p{0.5cm}|p{2cm}|p{0.5cm}|p{2cm}|p{0.5cm}|p{2cm}|}
\hline
IDs & Category & IDs  & Category & IDs  & Category  & IDs  & Category  & IDs  & Category \\
\hline
 a0 & Punching and knee lifting & a1 & Marking time and knee lifting & a2 & Jumping jack  &  a3 & Squatting & a4 & Forward lunging \\
a5 & Left lunging & a6 & Left stretching  & a7 & Raising hand and jumping  & a8 & Left kicking  & a9  & Rotation clapping \\
a10 & Front raising in turn & a11 & Pulling a chest expander  & a12 & Punching  & a13 & Wrist circling  & a14 & Single dumbbell raising \\
a15 & Shoulder raise & a16 & Elbow circling  & a17 & Dumbbell one-arm shoulder pressing  & a18 & Arm circling  & a19 & Dumbbell shrugging \\
a20 & Pinching back & a21 & Head anticlockwise circling  & a22 & Shoulder abduction  & a23 & Deltoid muscle stretching  & a24 & Straight forward flexion \\
a25 & Spinal stretching & a26 & Dumbbell side bend  & a27 & Standing opposite elbow-to-knee crunch  & a28 & Standing body rotation  & a29 & Overhead stretching \\
a30 & Upper back stretching & a31 & Knee to chest stretch & a32 & Knee circling  & a33 & Alternate knee lifting  & a34 & Bent over twist \\
a35 & Rope skipping & a36 & Standing toe touches & a37 & Standing Gastrocnemius Calf Stretch & a38 & Single-leg lateral hopping  & a39 & High knees running \\
\hline
\end{tabular}
\end{center}
\end{table*}

\textbf{Viewpoints} For the arbitrary-view action recognition, the dataset is designed to have more viewpoints and covers a wider range of view angles. Our dataset has 8 fixed viewpoints which averagely distribute around subjects, as shown in Figure~\ref{fig:captureSet}.
In order to simulate a real scenario concerning the HRI, we design three capture settings. The setting A in Figure~\ref{fig:captureSetA} is a circle with a radius of $2.5 m$, and there are two settings having the rectangle shape in Figure~\ref{fig:captureSetB}. The size of setting B and C are shown with black dashes and green dashes, respectively. In addition, we capture varying-view action sequences by moving a sensor around the subject along blue paths in the Figure~\ref{fig:captureSet}. It captures actions in continuously varying views covering $360^\circ$ full-circle view angles. We set the height of all Kinect sensors to $1.2 m$. The capture in three settings involves various camera-to-subject distances. Varying-view sequences are mainly used to evaluate approaches for arbitrary-view action recognition.

\textbf{Subjects} We totally invite 118 persons to attend the dataset collection. Each person averagely acts 10 action categories, and each action category is performed by 40 subjects in total. Because action categories are fitness actions, the age of subjects is from 18 to 30. We provide each subject a number ID in the collected dataset.

\begin{figure*}
 \begin{center}
 \subfigure[Frame samples of 13 action categories in 8 fixed viewpoints and varying-view sequences.]{
 \label{fig:DBexampleA} 
 \includegraphics[width=0.7\linewidth]{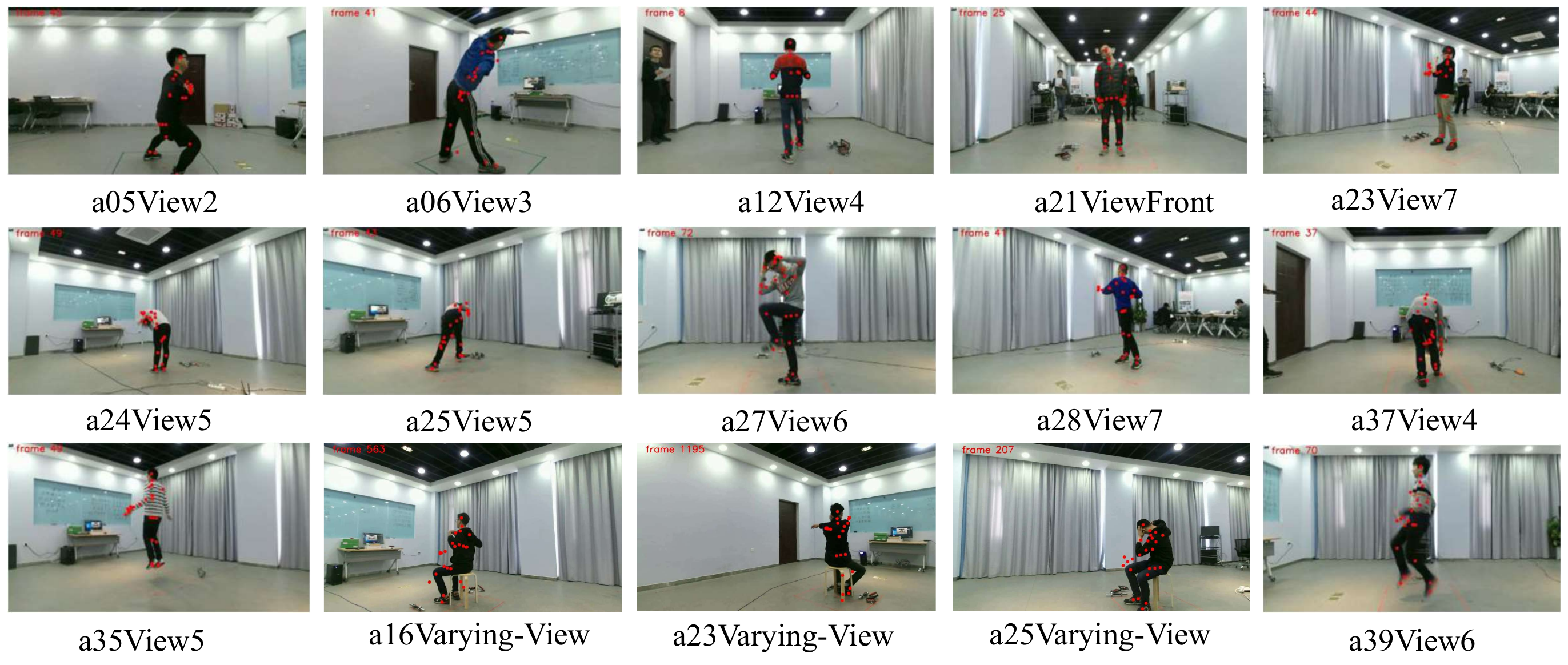}}
 \hspace{1in}
 \subfigure[Temporal frame samples in the varying-view sequence of action $a27$.]{
 \label{fig:DBexampleB} 
 \includegraphics[width=0.7\linewidth]{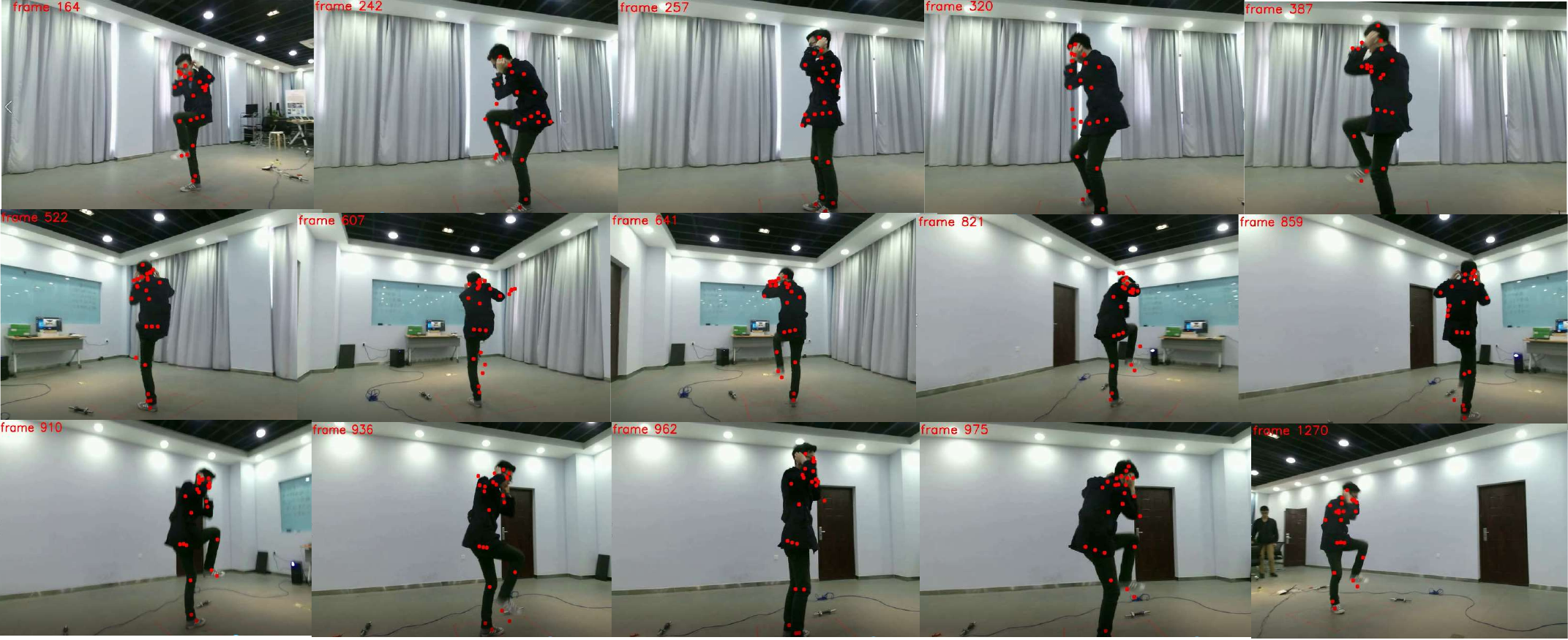}}
 \caption{ Frame samples in the varying-view RGB-D action dataset. }
 \label{fig:DBexample} 
 \end{center}
 \end{figure*}

\textbf{Categories} There are total of 40 action categories in the dataset. We show all the categories in the Table~\ref{tab:category}.
Among 40 categories, 15 categories are performed in two situations, standing to act and sitting on a chair to act. Other 25 action categories are all performed with the standing pose. These categories are given action IDs of $a0-a39$. These categories of actions contain both of large motions of all body parts, ~\eg spinal stretch, raising hands and jumping, and small movements of one part, ~\eg head anticlockwise circle. They are much more complex than actions in existing datasets,~\eg, hand waving, walking~\etc. Figure~\ref{fig:DBexampleA} shows frame samples of 13 action categories captured in 8 fixed viewpoints and in varying-view sequences, while Figure~\ref{fig:DBexampleB} displays temporal frames of the action $a27$ (Standing opposite elbow-to-knee crunch) in a varying-view sequence. As shown in the figure, our dataset consists of complex motions and rapidly changing poses. Captured skeletons have distortions and loss of joints.

\textbf{Quantities} When capturing actions in fixed viewpoints, each subject repeats each action $3 \sim 5$ times. For each side-view action capturing, ~\ie view1 $\sim$ view7, there lays a sensor in the front view to capture synchronous action sequences. We use three Kinect sensors to synchronously capture two side-view and one front-view actions. Similarly, when we capture varying-view action sequences, we synchronously capture the front-view sequences. Thus each side-view action sequence has a synchronous sequence in the front viewpoint. Totally, 11 sequences are captured in 8 fixed viewpoints and 2 varying-view sequences are recorded per action category per subject. These synchronous action sequences can be used for view transformation between side views and the front viewpoints. In our dataset, one RGB video in fixed viewpoints generally sustains about $6 \sim 7$ seconds, and contains $200+$ frames. RGB videos of varying-view sequences have about $55 \sim 65$ seconds, containing about $1730 \sim 2000$ frames. The length of RGB videos in our dataset are more than 83 hours in total. Depth and skeleton sequences have synchronization with RGB videos, thus depth sequences has similar frame numbers with the RGB videos.

\subsection{Comparison with other datasets}
We compare our action dataset with other multi-view action datasets in Table~\ref{tab:databaseCom}. It shows that our dataset has more subjects and viewpoints. Various subjects may be used to evaluate the generalization of recognition approaches with different persons. In terms of the viewpoint, besides of 8 fixed viewpoints, our dataset captures varying-view sequences which cover the $360^\circ$ full-circle views, which is superior to other datasets. Varying-view action sequences provide experiment samples for arbitrary-view action recognition.

Our dataset collects fitness actions because they involve both large movements and small motions, and it may be applied to fitness auxiliary with robots.
Referring to the sample quantity, our dataset contains a large-scale action samples. The sample number of RGB videos is 25,600, and total 72,709 samples of RGB videos, depth and skeleton sequences. More importantly, each varying-view sequence is ten times the length of sequences in other datasets. We estimate the total length of RGB videos in hours for existing multi-view datasets and list them in the Table~\ref{tab:databaseCom}. The comparison indicates that our dataset has the longest video playing hours.

\begin{table*}[!t] \footnotesize
\begin{center}
\caption{Comparison with other multi-view action datasets.}
\label{tab:databaseCom}
\begin{tabular}{|p{3cm}|p{1cm}|p{1.2cm}|p{2cm}|p{2cm}|p{2.8cm}|p{3cm}|}
\hline
Databases & Subjects & Categories & Viewpoints & Sensors & Data & Quantity (samples, RGB video length in hours)\\
\hline
IXMAS~\cite{IXMAS2006} & 10  & 11 & 5 & Camera & RGB & 550, $ < 2 $\\
$Act4^2$ ~\cite{ChengACT42012} & 24 & 14 & 4 & Kinect v1 & RGB,Depth & 6844, 34 \\
Multiview 3D Event~\cite{Wei4DHOI2013} & 8 & 8 & 3 & Kinect v1 & RGB,Depth,Skeleton & 3815, $\approx 3.5$ \\
Northwestern-UCLA~\cite{WangMSTAOG2014} & 10 & 10 & 3 & Kinect v1 & RGB,Depth,Skeleton & 1475, $ < 1$ \\
UWA3D Multiview~\cite{RahmaniUWA3D2016} & 10 & 30 & 5 & Kinect v1 & RGB,Depth,Skeleton & 1075, $\approx 1.5$ \\
NTURGB+D~\cite{ShahroudyNTU2016} & 40 & 60 & 5, 16 settings & Kinect v2 & RGB,Depth,Skeleton,IR  & 56,880, $\approx 79$ \\
\textbf{Ours} & 118 & 40 & 8 fixed + varying ($360^\circ$) & Kinect v2 & RGB,Depth,Skeleton & 25,600, 83 \\
\hline
\end{tabular}
\end{center}
\end{table*}

\subsection{Evaluations}
We evaluate approaches in our dataset using four types of evaluations. The standard evaluation of the cross-subject recognition in~\cite{ShahroudyNTU2016} is kept in our experiment. To evaluate the recognition performance between any two viewpoints and neighbor viewpoints, we propose two types of cross-view recognition. Furthermore, we evaluate the arbitrary-view action recognition in our dataset.

\textbf{Cross-subject recognition} In our dataset, each subject acts 10 actions, and each action is surely acted by 40 subjects. For cross-subject recognition, we randomly select 51 subjects, and separate action samples acted by these subjects into the training group. The group consists of all action categories. The subject IDs selected for training are 1, 2, 6, 12, 13, 16, 21, 24, 28-31, 33, 35, 39, 41, 42, 45, 47, 50, 52, 54, 55, 57, 59, 61, 63, 64, 67, 69-71, 73, 77, 81, 84, 86-88, 90, 91, 93, 96, 99, 102-104, 107, 108, 112, 113.
Action samples of the rest subjects are put into testing groups. The separation rule is used in all cross-subject recognition experiments in this paper.

\textbf{Cross-view recognition I} To evaluate the performance of action recognition in cross viewpoints, action samples in one of 8 fixed viewpoints are used for training, and the test is operated on samples in another fixed viewpoint. Results of the cross-view recognition are reported using a confusion matrix of all viewpoints.

\textbf{Cross-view recognition II} The evaluation is in order to show the recognition performance between neighbor viewpoints. Action samples in four viewpoints which connect a square crossing shape in the Figure~\ref{fig:captureSet} are used as training samples, and samples in the other four viewpoints are regarded as test samples. For example, samples of the front viewpoint, viewpoints 4, 2 and 6 are separated into a training group, and samples of viewpoints 1, 3, 5, and 7 are separated into the test group. Conversely, training is operated using samples captured in viewpoints 1, 3, 5 and 7, and samples of the front viewpoint, viewpoints 4, 2 and 6 are used for the test.

\textbf{Arbitrary-view recognition} We evaluate arbitrary-view recognition in two ways. In a first way (Arbitrary-view I), we use action samples captured in 8 fixed viewpoints to train classifier models, and evaluate trained models on varying-view sequences. In the other way (Arbitrary-view II), we use varying-view sequences for training and also test the trained model on varying-view sequences. Action sequences captured in continuously varying views generally contain 2000+ frames per sequence. The sequence length is 10 times of action sequences captured in fixed viewpoints. In order to do experiments as the same situation as other evaluations, we clip each varying-view action sequence to 10 short clips with an equal length so that each short section has the similar length with sequences captured in fixed views. These short sections are used independently in the evaluation of recognition models.

\section{View-guided Skeleton-CNN}
\label{sec:VSK-CNN}

\subsection{The VS-CNN network}
\begin{figure}[t]
\begin{center}
\includegraphics[width=1\linewidth]{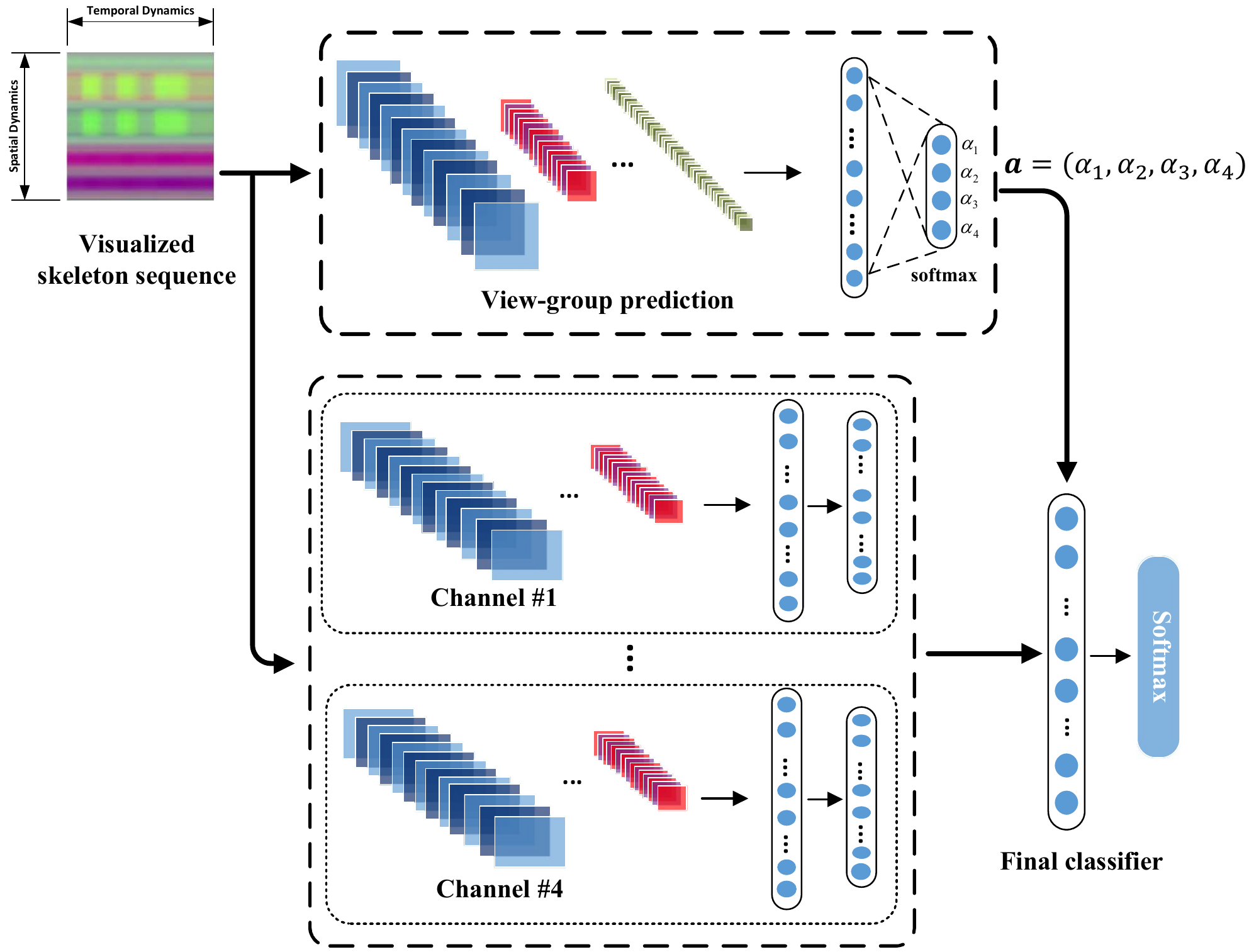}
\end{center}
   \caption{The architecture of the VS-CNN. The view-group predictor separates samples into four view groups and guides the training of corresponding feature learning channels and classifiers. Finally, a weighted fusion is performed on output score features of four classifiers and to train a classifier for final recognition. }
\label{fig:VSCNN}
\end{figure}

The architecture of the VS-CNN is shown in Figure~\ref{fig:VSCNN}. Since skeleton visualization~\cite{EnhancedSK2017} is able to somewhat weaken difference of features in various views, we employ it to generate an initial skeleton representation of actions and use the representation as input features of VS-CNN. Moreover, as action samples in our dataset cover the $360^\circ$ full-circle view, it is difficult to use one traditional feature learning model to learn correct feature representations for all views. Thus, we separate the full-circle view into four view groups and design four feature learning channels which correspond to four view groups. A view-group predictor is designed to determine view groups for action samples and guides the training of corresponding feature learning channels and classifiers by inputting samples to corresponding channels according to prediction result of the view-group predictor. Then we fuse output score features of four channels and train a classifier to finally determine action categories.

\begin{figure}[t]
\begin{center}
\includegraphics[width=0.45\linewidth]{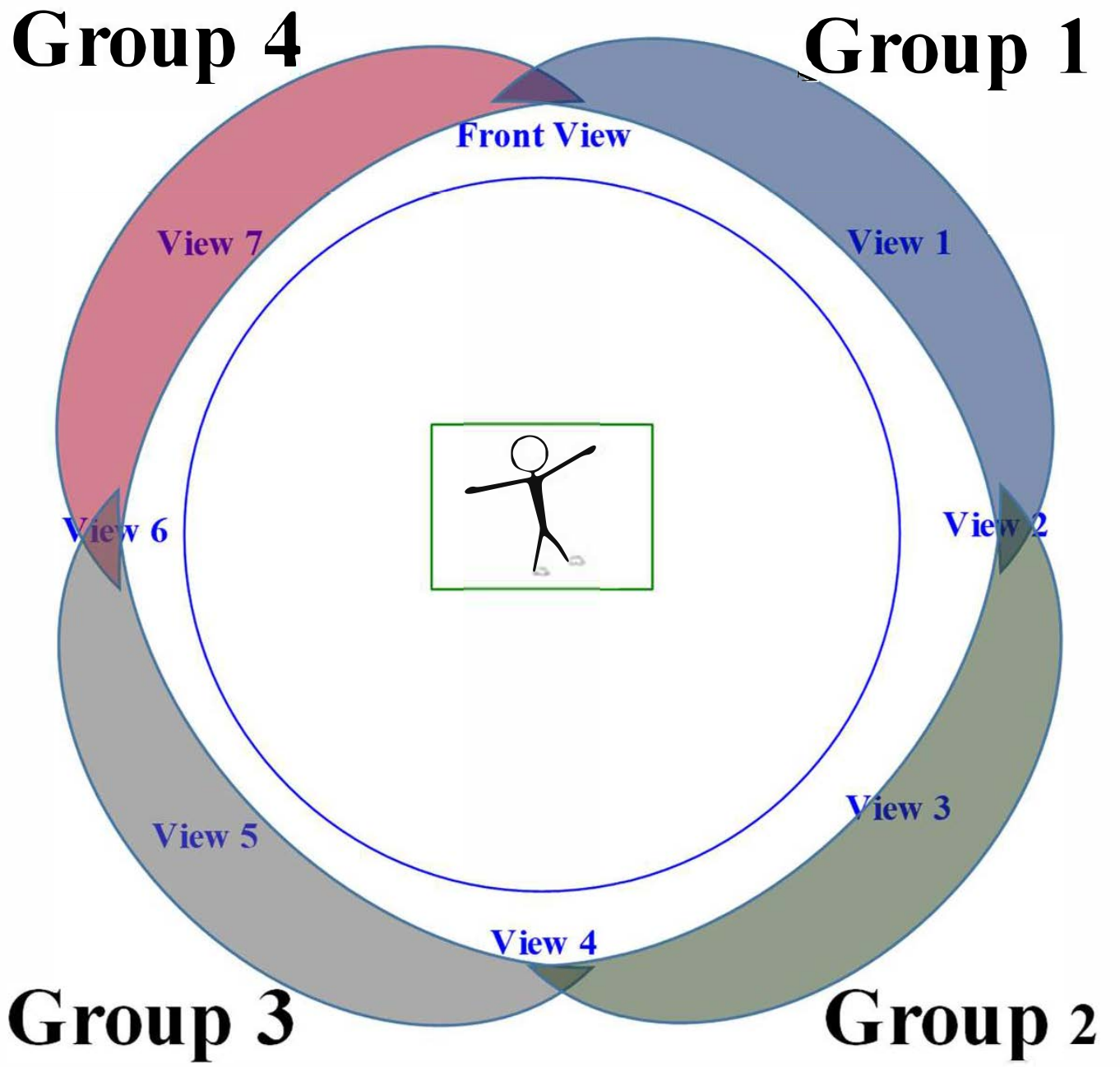}
\end{center}
   \caption{The $360^\circ$ full-circle view is separated into four groups, and these groups overlap each other. These overlaps will guide the VS-CNN to learn a common action representation in different view groups.}
\label{fig:viewGroup}
\end{figure}
\subsubsection{View-group predictor}
We separate the full-circle view into four view groups, and design a view-group predictor to realize automatically separation of action samples.
View group separation is shown in Figure~\ref{fig:viewGroup}. The front viewpoint, viewpoints 1, 2 are separated into the first view group, while viewpoints 2, 3, 4 are defined as the second view group. Similarly, the third view group includes viewpoints 4, 5, and 6, and the fourth group includes viewpoints 6, 7, and the front viewpoint. These groups overlap each other so that samples belonging to overlap views drive the training of two feature learning channels. Since any two feature learning channels share part of common samples during training, each channel learns a common representation for samples in neighbor view groups. In this way, we obtain a common representation of action samples in a full-circle view range. Therefore, our approach is able to overcome view invariance.

The view-group predictor consists of 3 CNN layers, and a SoftMax is employed as classifier to determine the probability of one sample belonging to four view groups. The structure of the CNN network is: layer 1 ( 16 kernels, kernel size $3\times3$, 1 stride), layer 2 ( 32 kernels, kernel size $3\times3$, 1 stride ), layer 3 ( 32 kernels, kernel size $3 \times 3$, 2 strides ). Suppose that $x$ represents an action sample, we use function $f_g(\bm{\theta}_z, x)$ to represent the group prediction network, and $\bm{\theta}_z$ refers to network parameters. The view-group predictor outputs a vector, $\hat{\textbf{y}}_g = \{ z_i, i=1,\cdots, 4\}$, which indicates prediction scores of four view groups. Here, $i$ refers to the $i$th view group. The prediction score is further used to calculate weights by Equation~(\ref{eqn:softmax}), $\bm{a}= \{\alpha_i, i=1,\cdots, 4\}$, where $\sum{\alpha_i} =1$. Weights $\bm{a}$ are used for the weighted fusion of score features output from four classifiers. Equation~(\ref{eqn:lossGroup}) is designed as the prediction loss to train the view-group prediction network. Here, $\textbf{y}_g$ refers to the ground truth of view groups. $\omega$ refers to parameters of the view-group predictor.

\begin{equation}
\begin{aligned}
 \alpha_i = \frac{e^{-z_i}}{\sum\limits_{i=1}^4{e^{-z_i}}}.
\label{eqn:softmax}
\end{aligned}
\end{equation}

\begin{equation}
\begin{aligned}
\hat{\textbf{y}}_g & = \frac{\exp( \omega f_g(\bm{\theta}_z, x) } {\sum_i \exp( \omega f_g(\bm{\theta}_z, x) ) }. \\
L_g( x,\textbf{y}_g ) & = -\textbf{y}_g \log \hat{\textbf{y}}_g. \\
\label{eqn:lossGroup}
\end{aligned}
\end{equation}

\subsubsection{View-guided feature learning channels}
Corresponding to four view groups, we design four feature learning channels using the SK-CNN~\cite{EnhancedSK2017} as base network. Each feature learning channel is composed of an SK-CNN backbone and one action classifier.
According to $\max(z_i)$, action samples are separated into four view groups, and they are input to corresponding feature learning channels to learn action features and the following classifier gives a prediction score vector of action categories. Suppose that $x$ is an action sample, we use $f_i( \bm{\theta}_i, x)$ to represent feature learning of the $i$th channel. Here, $\bm{\theta}_i$ is network parameter.
We employ a Softmax classifier to predict action categories of action samples. The cross entropy is adopted as a loss function for the training of feature learning networks and Softmax classifiers. The Equation~(\ref{eqn:lossChannel}) shows a loss function of the $i$th channel. $\omega_i$ refers to classifier parameter. Here, $\textbf{y}$ refers to the ground truth of action categories, and $\hat{\textbf{y}}^i$ represents predicted results, $\hat{\textbf{y}}^i = \{ \hat{y}^i_j, j=1, \cdots, 40 \}$. $j$ refers to action category.

\begin{equation}
\begin{aligned}
\hat{\textbf{y}}^i & = \frac{\exp( \omega_i f_i( \bm{\theta}_i, x) ) } {\sum_j \exp( \omega_i f_i( \bm{\theta}_i, x) ) }. \\
L_i(x,\hat{\textbf{y}}^i) & = -\textbf{y} \log \hat{\textbf{y}}^i. \\
\label{eqn:lossChannel}
\end{aligned}
\end{equation}

\subsubsection{Channel fusion recognition}
With action prediction scores output from four channels, a weighted fusion is performed through a fully connected layer with 40 neurons. Following that, a SoftMax classifier is adopted for the final action category determination. We also use the cross-entropy as loss function for classifier training, as shown in Equation~(\ref{eqn:lossClass}). Here, $\textbf{y}$ refers to the ground truth of action categories, and $\hat{\textbf{y}}$ represents predicted results of action categories, $\hat{\textbf{y}} = \{ \hat{y}_j, j=1, \cdots, 40 \}$. $\omega_j$ is classifier parameter. Here, $\alpha_i, i=1,\cdots, 4$ is used to weight prediction scores of four channels.

\begin{equation}
\begin{aligned}
\hat{\textbf{y}} & = \frac{\exp( \omega_j \sum_{i=1}^4 {\alpha_i \hat{\textbf{y}}^i} ) } { \sum_j \exp( \omega_j \sum_{i=1}^4 {\alpha_i \hat{\textbf{y}}^i} ) }. \\
L(x,\textbf{y}) & = -\textbf{y} \log \hat{\textbf{y}}. \\
\label{eqn:lossClass}
\end{aligned}
\end{equation}

\subsection{Training and Testing}
\textbf{Training phase}
We employ the stochastic gradient descent algorithm to minimize loss functions, and train optimal parameters for VS-CNN. The training is performed in three steps. In the first step, we assign action samples with labels of view groups, and train the view-group predictor. In the second step, we input training samples to the trained view-group predictor and automatically separate samples into different view groups. These separated samples are given to corresponding feature learning channels to train feature learning networks and classifiers with the loss function of Equation~(\ref{eqn:lossChannel}). In the final step, we fuse prediction scores of four classifiers with weights $\alpha_i$ and perform an end-to-end training again on the full VS-CNN network. The operation adjusts parameters in the view-group predictor and four feature learning channels, and seeks for optimal parameters for the final classifier.
In the experiment of cross-view recognition I, since only one viewpoint is used for training, not all four channels are necessary. Thus we set $\alpha_1 =1, \alpha_2 \sim \alpha_4 =0 $, to train one channel, ignoring other channels for network training.

\textbf{Testing phase}
For testing, one sample is input to the view-group prediction module, and the predictor generates view-group scores $\{ \widehat{z}\in \Re^{1\times 4} \}$. It is used to calculate weight $\{\widehat{\alpha}\in \Re^{1\times 4}\}$. Following that, the testing sample is input to four feature learning channels for feature learning and obtaining prediction scores of action categories via channel classifier. We fuse prediction results by assigning weights $\{\widehat{\alpha}\}$ to them, and input fused score feature to the final classifier to determine the final recognition category for the test sample.


In addition, we modify the skeleton visualization approach~\cite{EnhancedSK2017} by calculating the inter-frame difference of skeletons and adding the motion information to visualized skeleton images. The representation modification further weakens differences in different views so that it deals with the problem of view variety in action recognition.

\subsection{Analysis of weight parameters $\alpha_i$}
Based on the stochastic gradient descent algorithm, the parameter $\bm{\theta}_z$ and $\bm{\theta}_i$ in VS-CNN network is updated following the Equation~(\ref{eqn:updata}) in training processing.
\begin{equation} \footnotesize
\bm{\theta}:  = \bm{\theta} - \eta \frac{\partial L(x,y)}{\partial \bm{\theta}}.
\label{eqn:updata}
\end{equation}


%
%

Observing the equation~(\ref{eqn:lossClass}), $\alpha_i$ controls the parameter update of the $i$th network during training. When the $\alpha_i$ have a large value, parameter $\bm{\theta}_i$ will be updated. Otherwise, the classifier network parameter $\bm{\theta}_i$ will be updated slowly or not be updated. In other words, the $\alpha_i$ drives one of four classifiers to be trained during the training. It means that the view-group prediction module guides four classifier training in the VS-CNN. Therefore, the VS-CNN is able to classify action samples correctly in each view group and can deal with the full-circle-view action recognition by fusing four classifiers together.

\section{Experiments and Discussions}
\label{sec:experiment}

We evaluate existing approaches and our proposed approach on the newly collected dataset. Four types of evaluations are performed,~\ie the cross-subject recognition, the cross-view recognition I and II, and the arbitrary-view recognition.

Using RGB videos, we report evaluation results of the joint heterogeneous features learning (JOULE) model~\cite{JOULE2015,JOULE2016}, the ResNeXt network~\cite{ResNeXtARX2018}, C3D (Convolutional 3D Network)~\cite{C3DICCV2015} and LRCN (Long-term Recurrent Convolutional Network)~\cite{LRCNCVPR2015}. For the LRCN network, we use two feature learning networks, resnet34 and resnet50.
We also give the evaluation report of depth sequences using the C3D approach~\cite{C3DICCV2015}.
For skeleton data, we evaluate the Temporal Convolutional Neural Networks (TCN)~\cite{TCN2017} and its modified version Res-TCN~\cite{ResTCN2017}, LSTM with 3D rotated skeletons~\cite{ShahroudyNTU2016}, P-LSTM~\cite{ShahroudyNTU2016}, SK-CNN~\cite{EnhancedSK2017}, and the ST-GCN~\cite{STGCN2018} for four types of evaluations. The two-layer LSTM and P-LSTM in~\cite{ShahroudyNTU2016} are adopted for the evaluation.

For RGB videos and depth sequences, we evenly select 20 frames in each action sequence, and evenly select 40 frames in each skeleton sequence for experiments. The proposed VS-CNN model is evaluated on our dataset, and the results of four types of evaluations are compared with related approaches. In experiments, the average recognition accuracy is recorded for the comparison of performance. In the following experiment results, $ FV $ refers to the front viewpoint, and $V1 \sim V7$ refer to the viewpoint $1 \sim 7$.

\begin{table*} \footnotesize 
\begin{center}
\caption{Results of four types of evaluations. All approaches perform better in the Cross-subject recognition than other evaluations because training and test samples have the same viewpoints. The result of cross-view II is a little worse than the cross-subject recognition. }
\label{tab:resultAll}
\begin{tabular}{|p{1cm}|p{2.3cm}|p{2cm}|p{2cm}|p{2cm}|p{2cm}|p{2cm}|}
\hline
Source & Approach & Cross-subject & Cross-View I  & Cross-View II  & Arbitrary-view I & Arbitrary-view II \\
\hline
\multirow{5}{*}{RGB} & JOULE~\cite{JOULE2016} & 0.65 & 0.31 & 0.60 & 0.35 & 0.60\\
 & ResNeXt~\cite{ResNeXtARX2018} & 0.58 & 0.32 & 0.48 & 0.43 & 0.52\\
 & C3D~\cite{C3DICCV2015} & 0.37 & 0.24 & 0.38 & 0.43 & 0.48 \\
 & LRCN(Resnet34)~\cite{LRCNCVPR2015}  & 0.45  &  0.11   &  0.17  & 0.29   & 0.25  \\
 & LRCN(Resnet50)~\cite{LRCNCVPR2015} & 0.39  & 0.11 &  0.15  & 0.27   & 0.25  \\
\hline
\multirow{1}{*}{Depth} & C3D~\cite{C3DICCV2015} & 0.53 & 0.10 & 0.22 & 0.34 & 0.33 \\
\hline
\multirow{7}{*}{Skeleton} & TCN~\cite{TCN2017} & 0.56 & 0.16 & 0.43 & 0.41 & 0.64\\
 & Res-TCN~\cite{ResTCN2017} & 0.63 & 0.14 &  0.48 & 0.47 & 0.68\\
 & LSTM~\cite{ShahroudyNTU2016} & 0.56 & 0.16 & 0.31 & 0.48 & 0.68 \\
 & P-LSTM~\cite{ShahroudyNTU2016} & 0.60 & 0.13 & 0.33 & 0.47 & 0.50 \\
 & SK-CNN~\cite{EnhancedSK2017} & 0.59 & 0.26 & 0.68 & 0.43 & 0.77\\
 & ST-GCN~\cite{STGCN2018} & 0.71 & 0.25 &  0.56 & 0.53 & 0.43 \\
 & VS-CNN\textbf{(Ours)} & \textbf{0.76} & 0.29  & \textbf{0.71} & \textbf{0.57} & 0.75\\
\hline
\end{tabular}
\end{center}
\end{table*}

\subsection{Cross-subject recognition}
The evaluation uses action samples captured in 8 fixed viewpoints. From 118 subjects, we select 51 subjects and separate action samples acted by these subjects into the training group. We test action samples of rest subjects and record the statistic of the average recognition accuracy for each fixed viewpoint. Statistical results per viewpoint of all evaluated approaches are listed in Table~\ref{tab:crossSub}. The average result of all viewpoints for each approach is shown in Table~\ref{tab:resultAll}.

\begin{table*}[!t] \footnotesize
\begin{center}
\caption{Evaluation of cross-subject recognition. Accuracies of the viewpoint 1 $\sim$ 7 have a symmetrical distribution around the viewpoint 4 for all approaches. Viewpoints 3 and 5 have lower accuracies because of heavy occlusions.}
\label{tab:crossSub}
\begin{tabular}{|p{1 cm}|p{2.3cm}|p{1cm}|p{1cm}|p{1cm}|p{1cm}|p{1cm}|p{1cm}|p{1cm}|p{1cm}|}
\hline
Source & Approach & FV & V1 & V2 & V3  & V4 & V5 & V6 & V7  \\
\hline
\multirow{4}{*}{RGB} & JOULE~\cite{JOULE2016} & 0.84  & 0.74 & 0.53 & 0.59 & 0.60 & 0.57 & 0.55 & 0.78  \\
 & ResNeXt~\cite{ResNeXtARX2018} & 0.65  & 0.59 & 0.57 & 0.54 & 0.60 & 0.54 & 0.60 & 0.59 \\
 & C3D~\cite{C3DICCV2015} & 0.47  & 0.32  & 0.32  & 0.33  & 0.44  & 0.40  & 0.33  & 0.34   \\
 & LRCN(Resnet34)~\cite{LRCNCVPR2015} & 0.49  & 0.44  & 0.41  & 0.42  & 0.49  & 0.46  & 0.42  & 0.47  \\
 & LRCN(Resnet50)~\cite{LRCNCVPR2015} & 0.47  & 0.36  & 0.35  & 0.41  & 0.44  & 0.40  & 0.37  & 0.36  \\
 \hline
\multirow{1}{*}{Depth} & C3D~\cite{C3DICCV2015} & 0.54  &  0.53 & 0.53  & 0.53  & 0.52  & 0.53 & 0.51  & 0.56 \\
\hline
\multirow{7}{*}{Skeleton} & TCN~\cite{TCN2017} & 0.78 & 0.67 & 0.49 & 0.40 & 0.49 & 0.45 & 0.50 & 0.67 \\
 & Res-TCN~\cite{ResTCN2017} & 0.81 & 0.67 & 0.60 & 0.53 & 0.55 & 0.53 & 0.61 & 0.73  \\
 & LSTM~\cite{ShahroudyNTU2016} & 0.79 & 0.67 & 0.43 & 0.47 & 0.52 & 0.48 & 0.43 & 0.66  \\
 & P-LSTM~\cite{ShahroudyNTU2016} & 0.78 & 0.70 & 0.52 & 0.49 & 0.58 & 0.50 & 0.53 & 0.74 \\
 & SK-CNN~\cite{EnhancedSK2017} & 0.75 & 0.67 & 0.54 & 0.48 & 0.57 & 0.47 & 0.57 & 0.63 \\
 & ST-GCN~\cite{STGCN2018} & 0.80 & 0.72 & 0.68 & 0.65 & 0.67 & \textbf{0.69} & \textbf{0.72} & 0.75 \\
 & VS-CNN\textbf{(Ours)} & \textbf{0.88}  & \textbf{0.85} &  \textbf{0.73} & \textbf{0.67} & \textbf{0.74} & \textbf{0.69} & 0.70 & \textbf{0.83} \\
\hline
\end{tabular}
\end{center}
\end{table*}

Table~\ref{tab:crossSub} shows that accuracies of the viewpoint 1 $\sim$ 7 have a symmetrical distribution around the viewpoint 4 for all approaches. Accuracies of viewpoints 3 and 5 are generally lower than other viewpoints because actions in the two viewpoints suffer heavy occlusions. The front viewpoint always gets the highest accuracy. Comparing results of the RGB video, depth sequence, and skeleton data, results obtained using skeleton data are better than the depth and RGB video, that declares the superiority of skeleton data in action recognition. Comparing the results of the RGB video and the depth sequence, the two action feature modalities have balance performance.
According to the Table~\ref{tab:resultAll}, compared with other approaches, the VS-CNN outperforms other methods. Except for the VS-CNN, the JOULE, the ST-GCN, and the Res-TCN also have good performance due to their robust ability of action feature representation.

\begin{figure}[t]
\begin{center}
\includegraphics[width=1\linewidth]{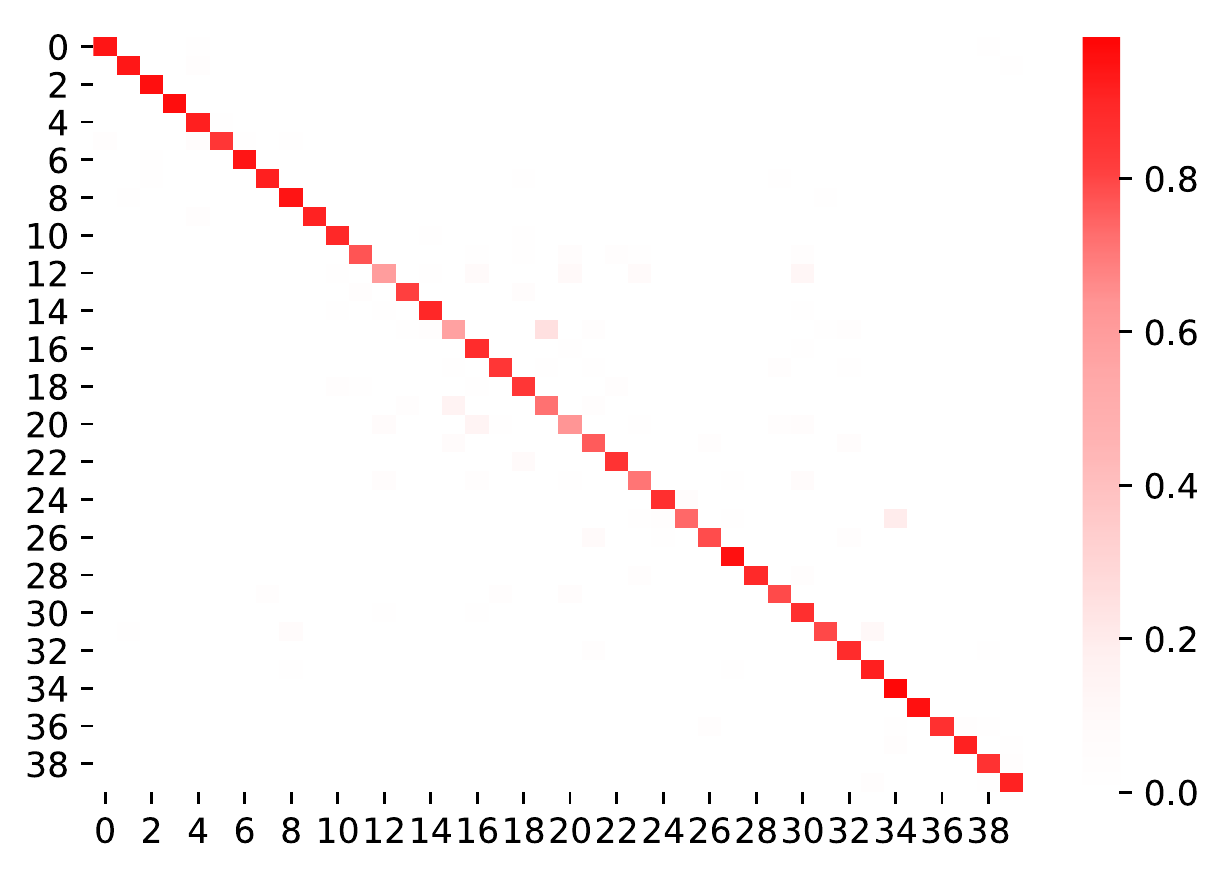}
\end{center}
   \caption{The confusion matrix of recognition results of cross-subject evaluation using the VS-CNN approach. It declares that actions in our dataset are much different from each other, so that is suitable for algorithm evaluation.}
\label{fig:confusionMatrix}
\end{figure}

In addition, we show the confusion matrix of recognition results of cross-subject evaluation using the VS-CNN approach in Fig.~\ref{fig:confusionMatrix}. Here, only skeleton data is used for the evaluation. The write color represents recognition results with a value of 0, and the red color represents recognition results of 1. As shown in the figure, most action categories have weak confusion with other categories. Thus the dataset provides suitable categories for algorithm evaluation.
\subsection{Cross-view recognition}
\begin{figure*}
 \centering
 \subfigure[JOULE.]{
 \label{fig:CrossJOULE} 
 \includegraphics[width=0.15\linewidth]{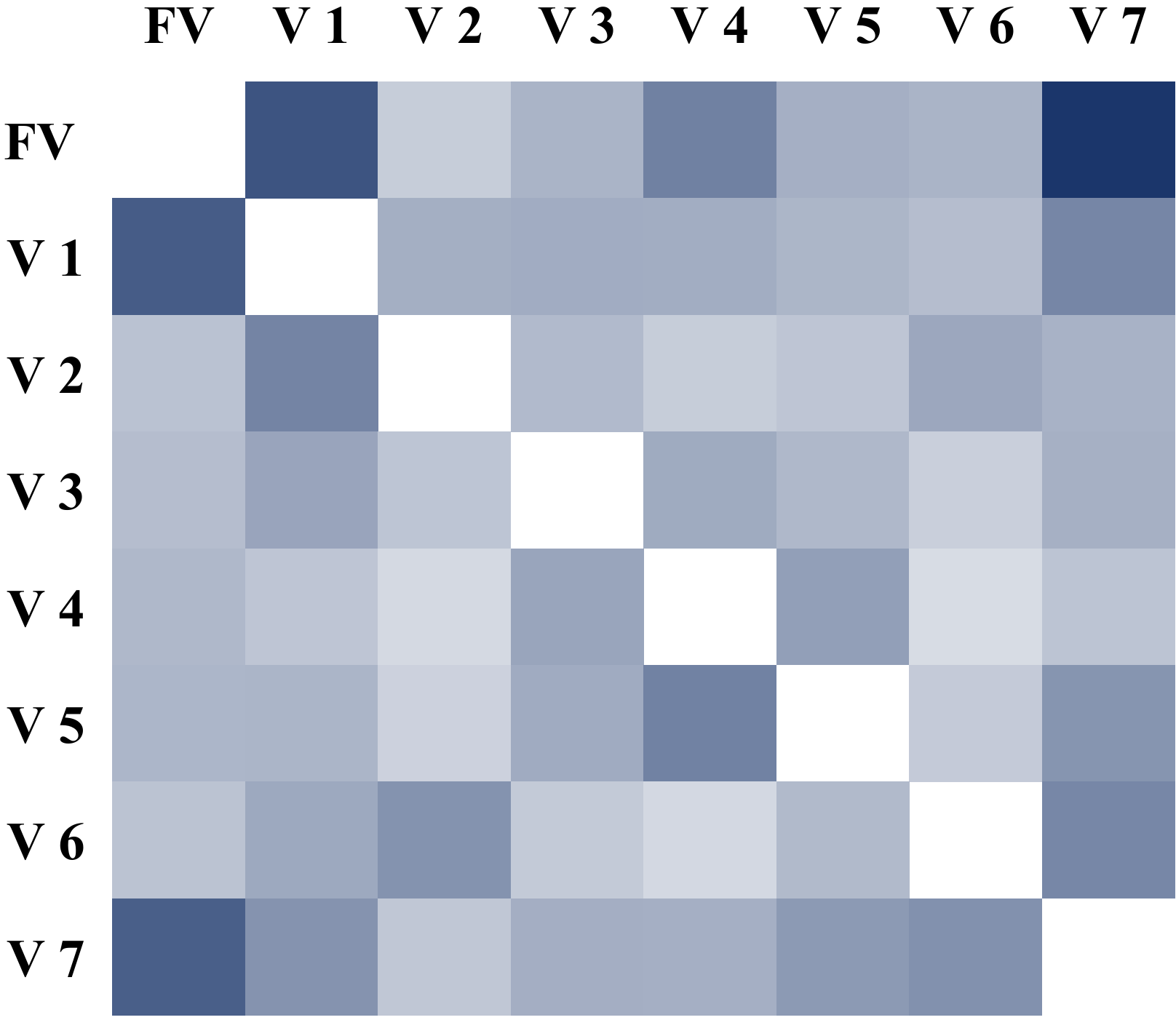}}
 \subfigure[ResNeXt.]{
 \label{fig:CrossResNeXt} 
 \includegraphics[width=0.15\linewidth]{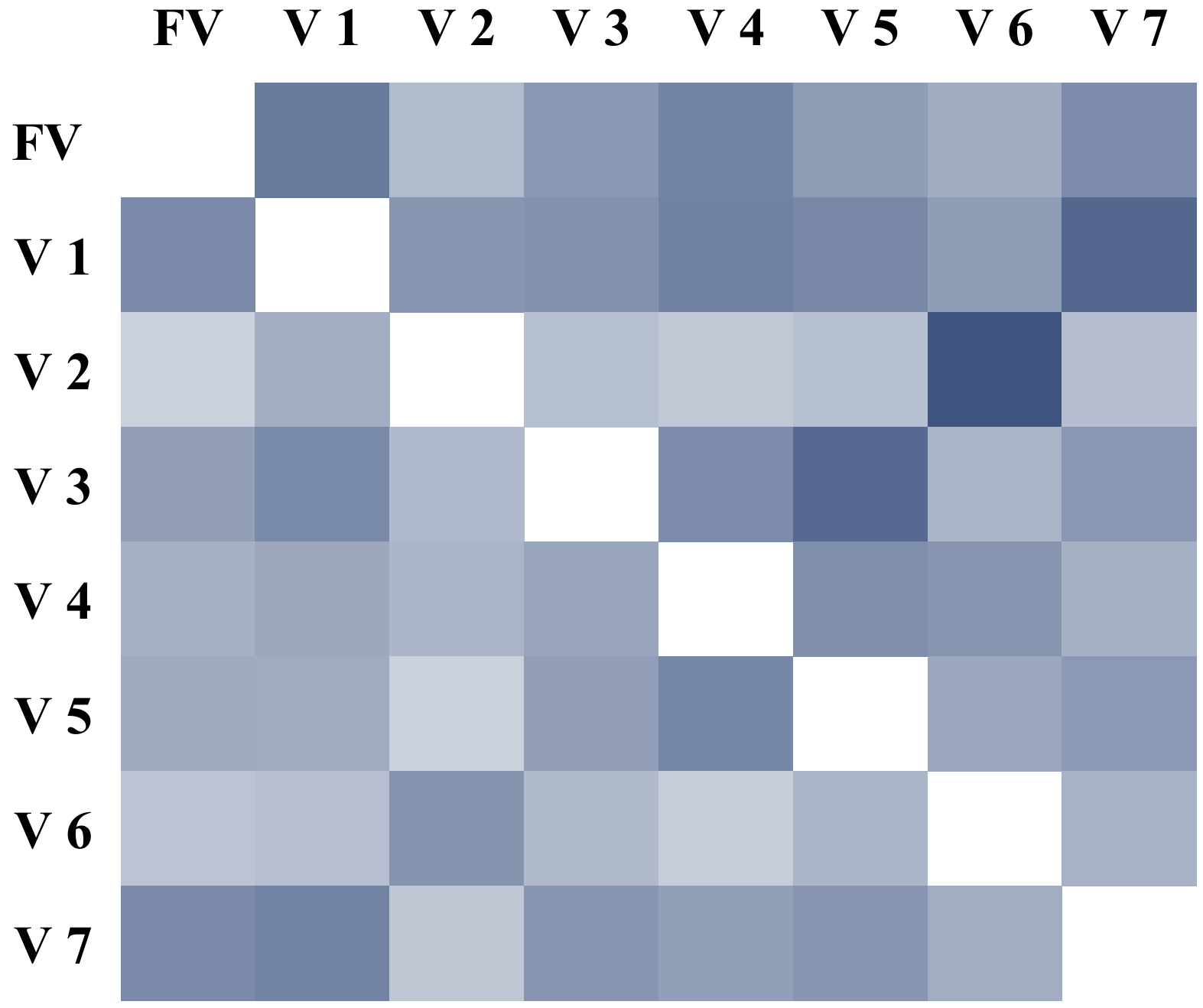}}
 \subfigure[TCN.]{
 \label{fig:CrossTCN} 
 \includegraphics[width=0.15\linewidth]{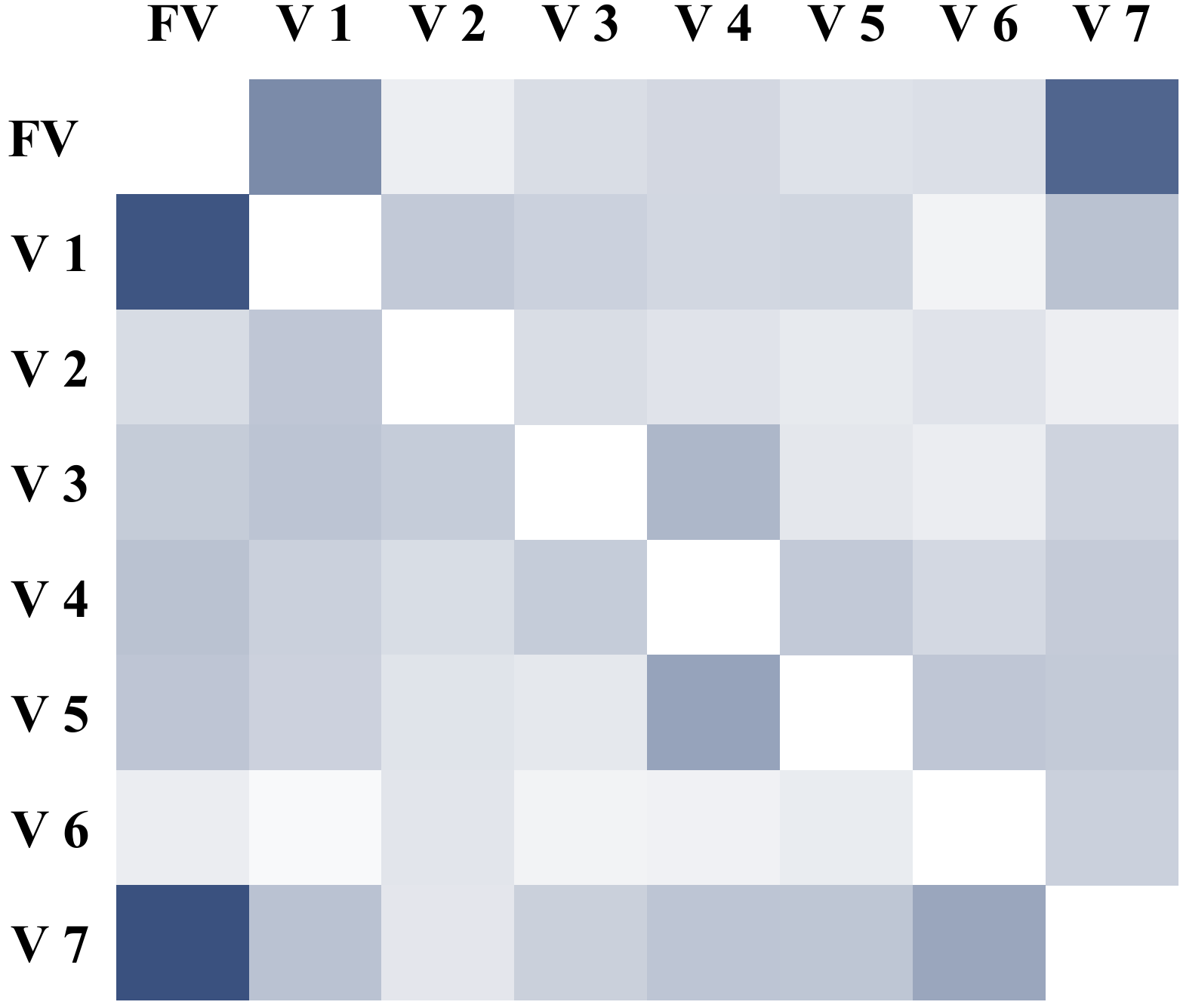}}
 \subfigure[Res-TCN.]{
 \label{fig:CrossRes-TCN} 
 \includegraphics[width=0.15\linewidth]{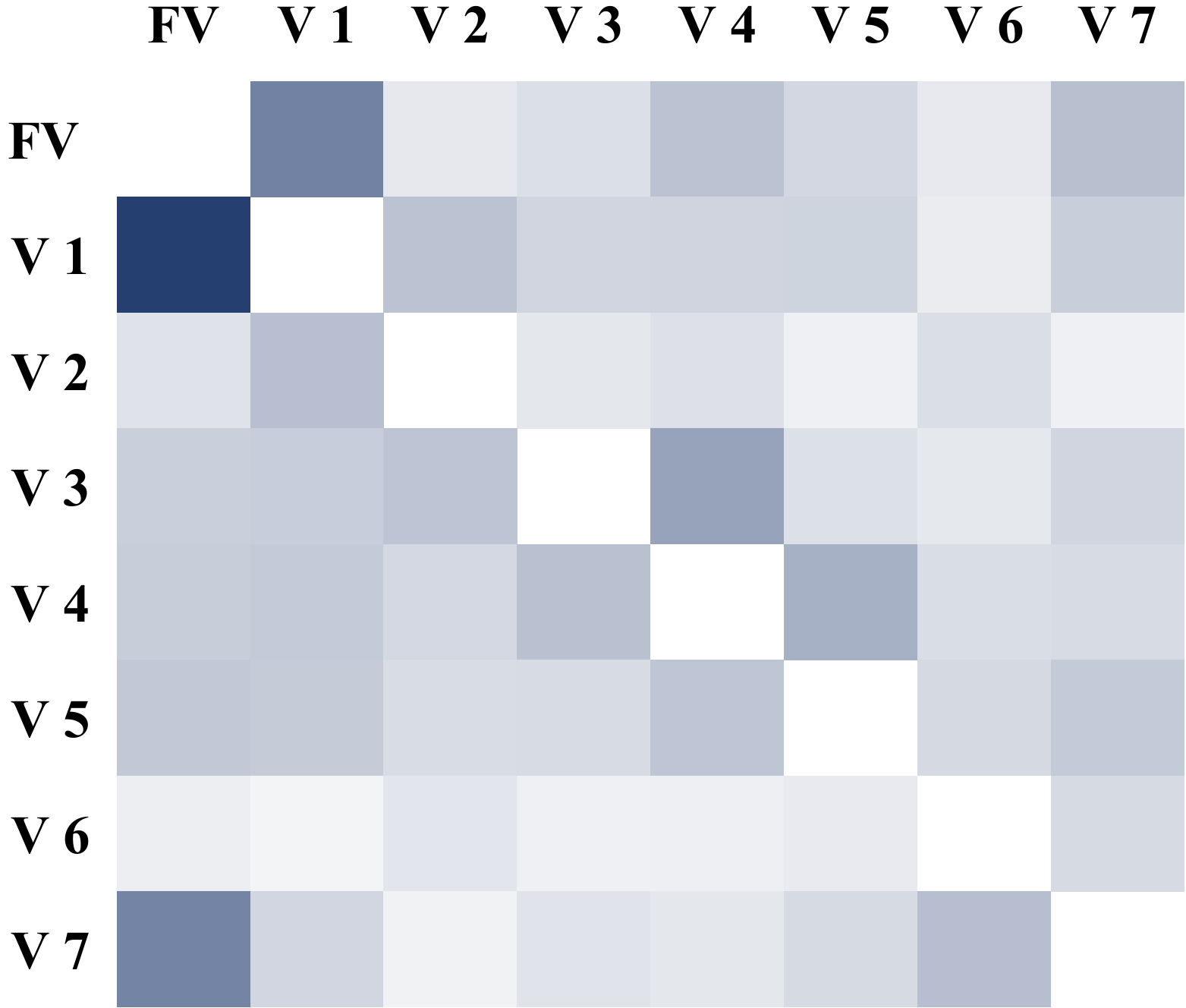}}
 \subfigure[LSTM.]{
 \label{fig:CrossLSTM} 
 \includegraphics[width=0.15\linewidth]{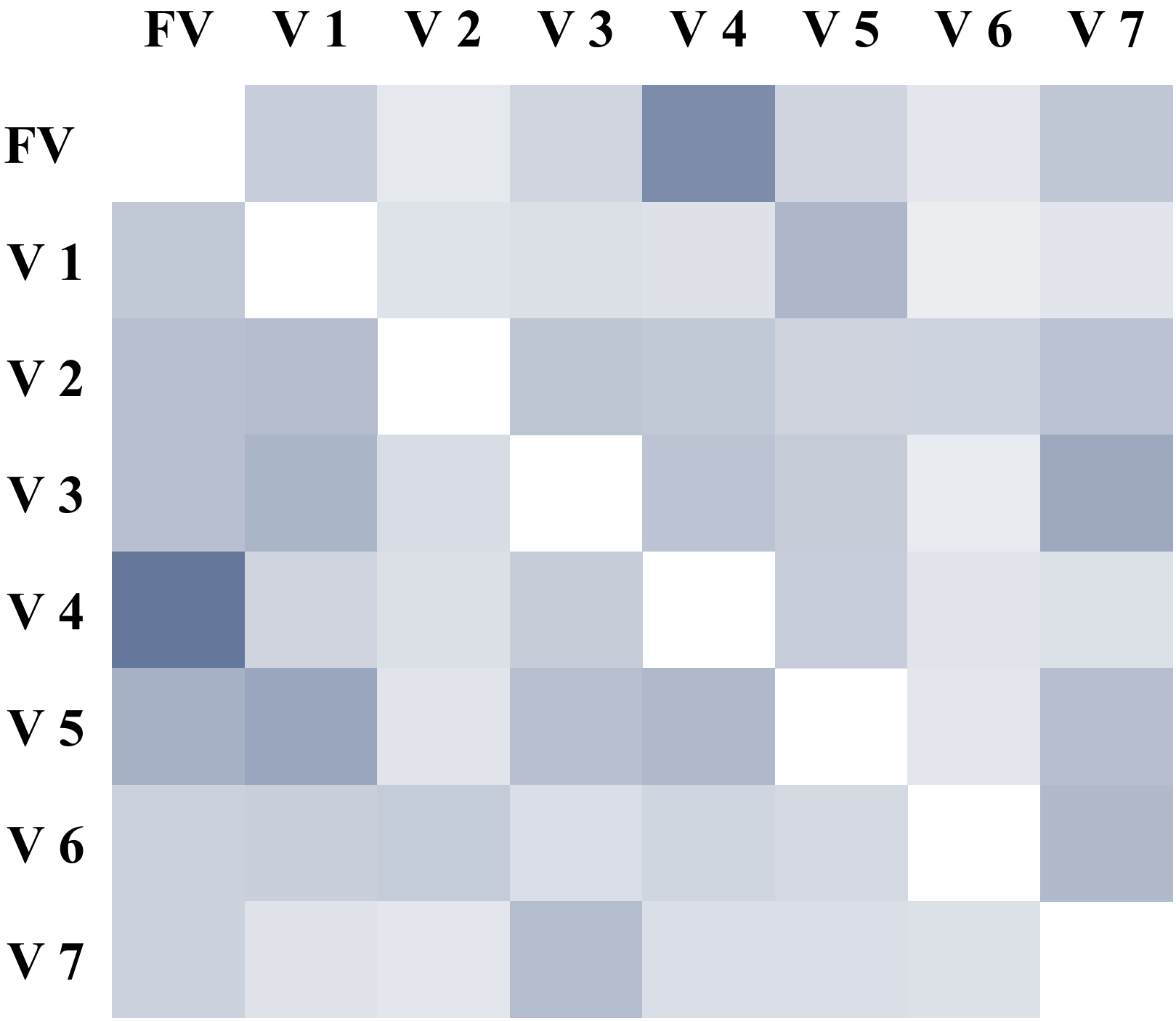}}
 \hspace{1in}
 \subfigure[P-LSTM.]{
 \label{fig:CrossP-LSTM} 
 \includegraphics[width=0.15\linewidth]{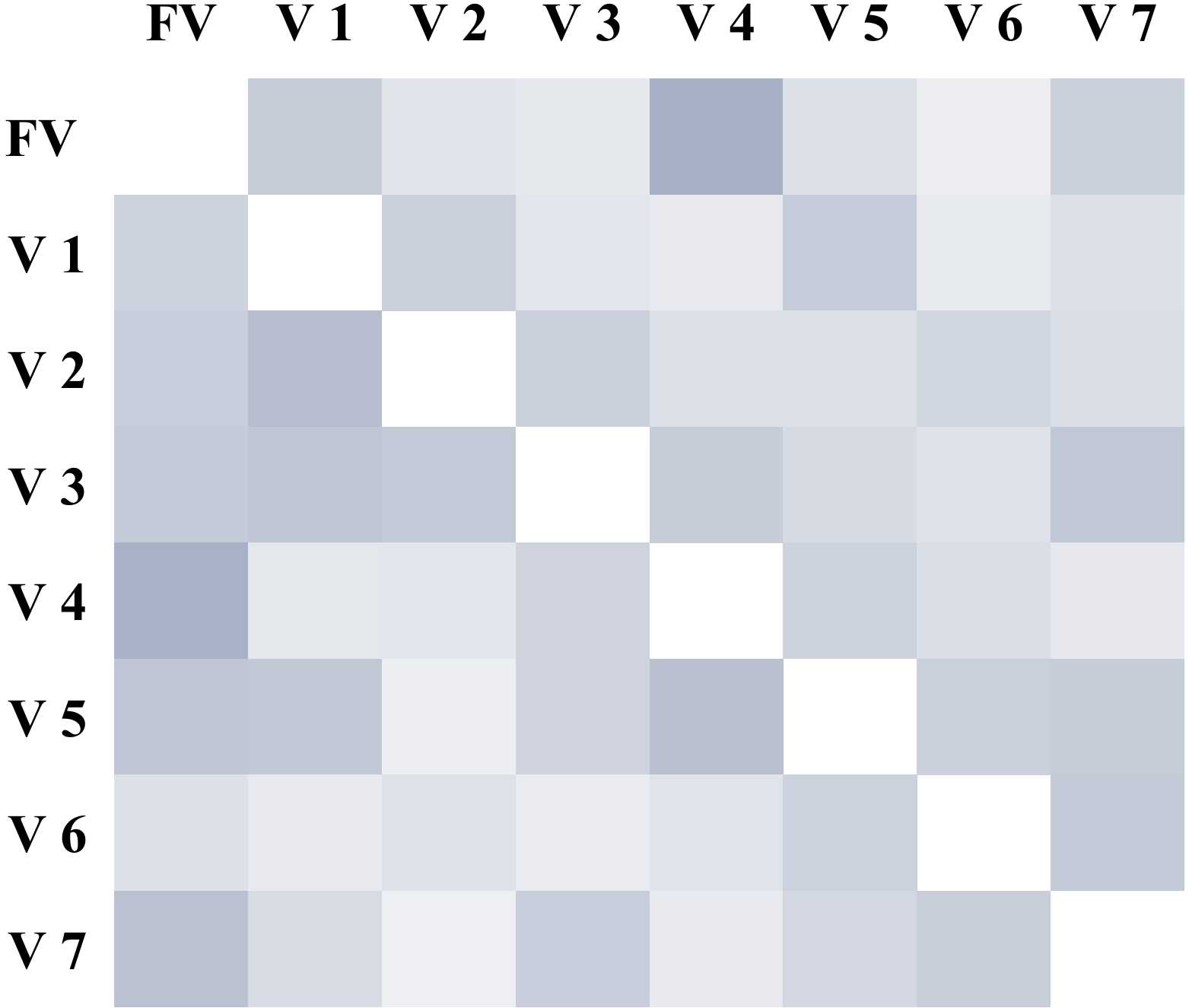}}
 \subfigure[Skeleton-CNN.]{
 \label{fig:CrossSkeleton-CNN} 
 \includegraphics[width=0.15\linewidth]{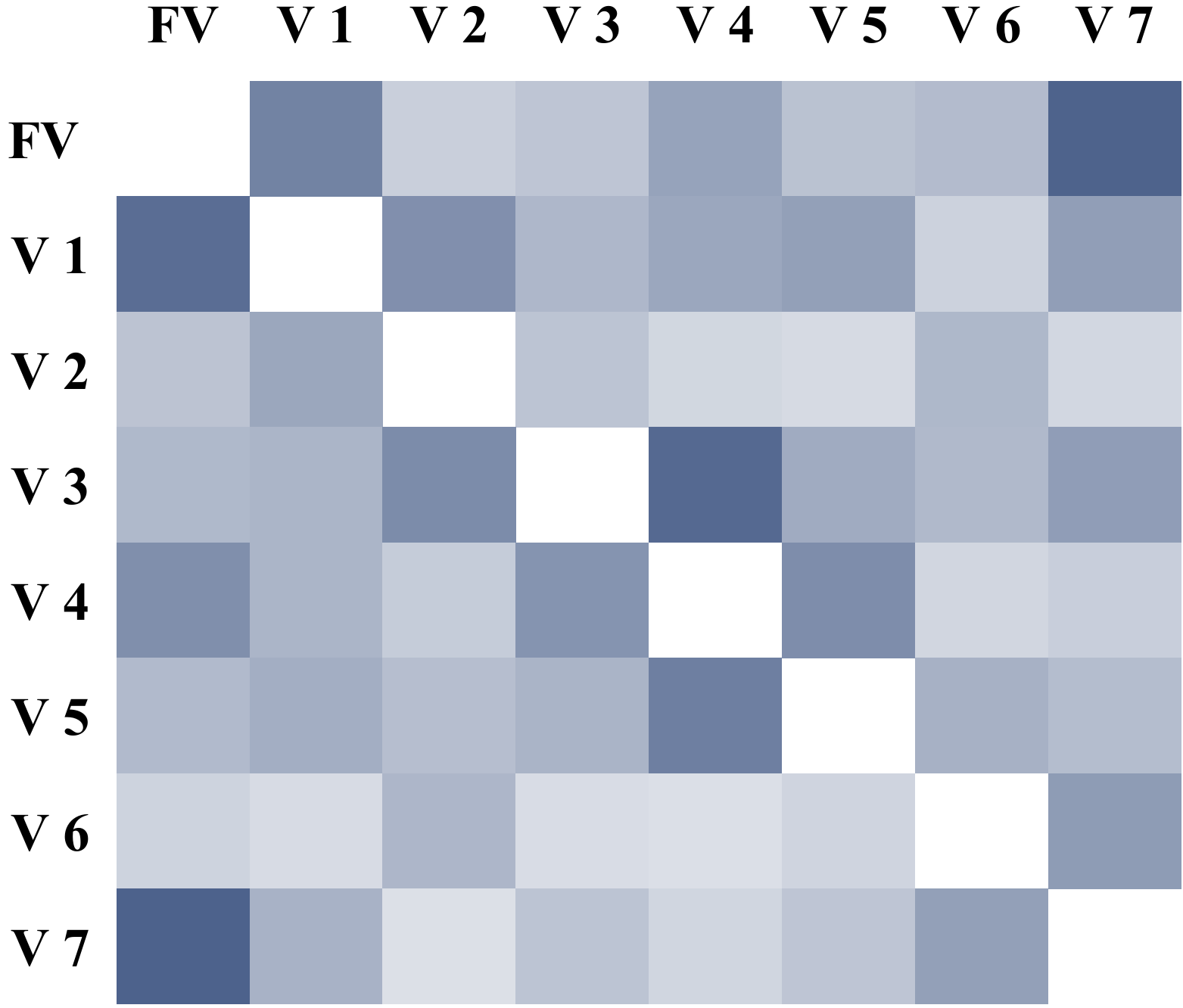}}
 \subfigure[ST-GCN.]{
 \label{fig:CrossST-GCN} 
 \includegraphics[width=0.15\linewidth]{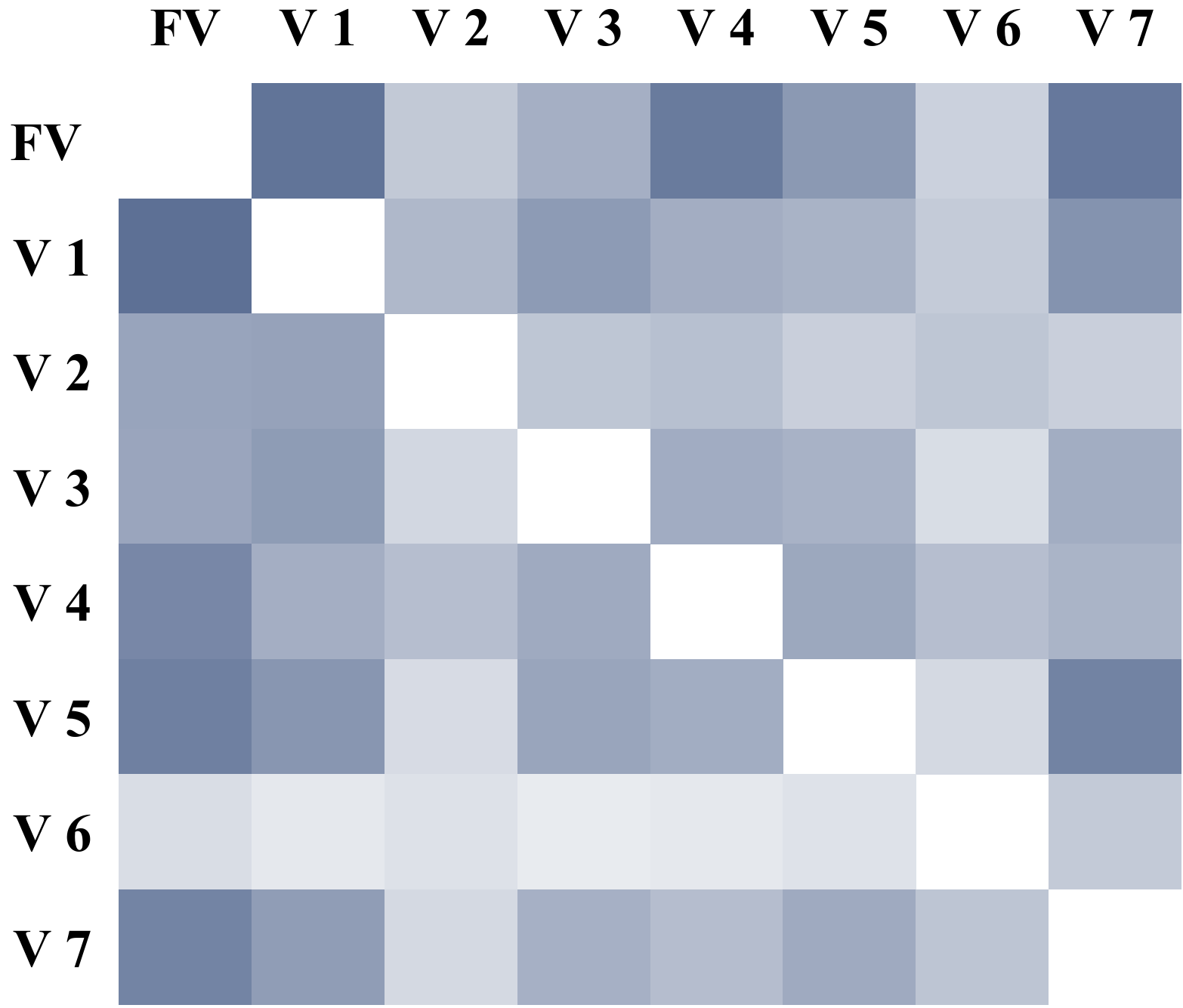}}
 \subfigure[VS-CNN.]{
 \label{fig:CrossP-LSTM} 
 \includegraphics[width=0.18\linewidth]{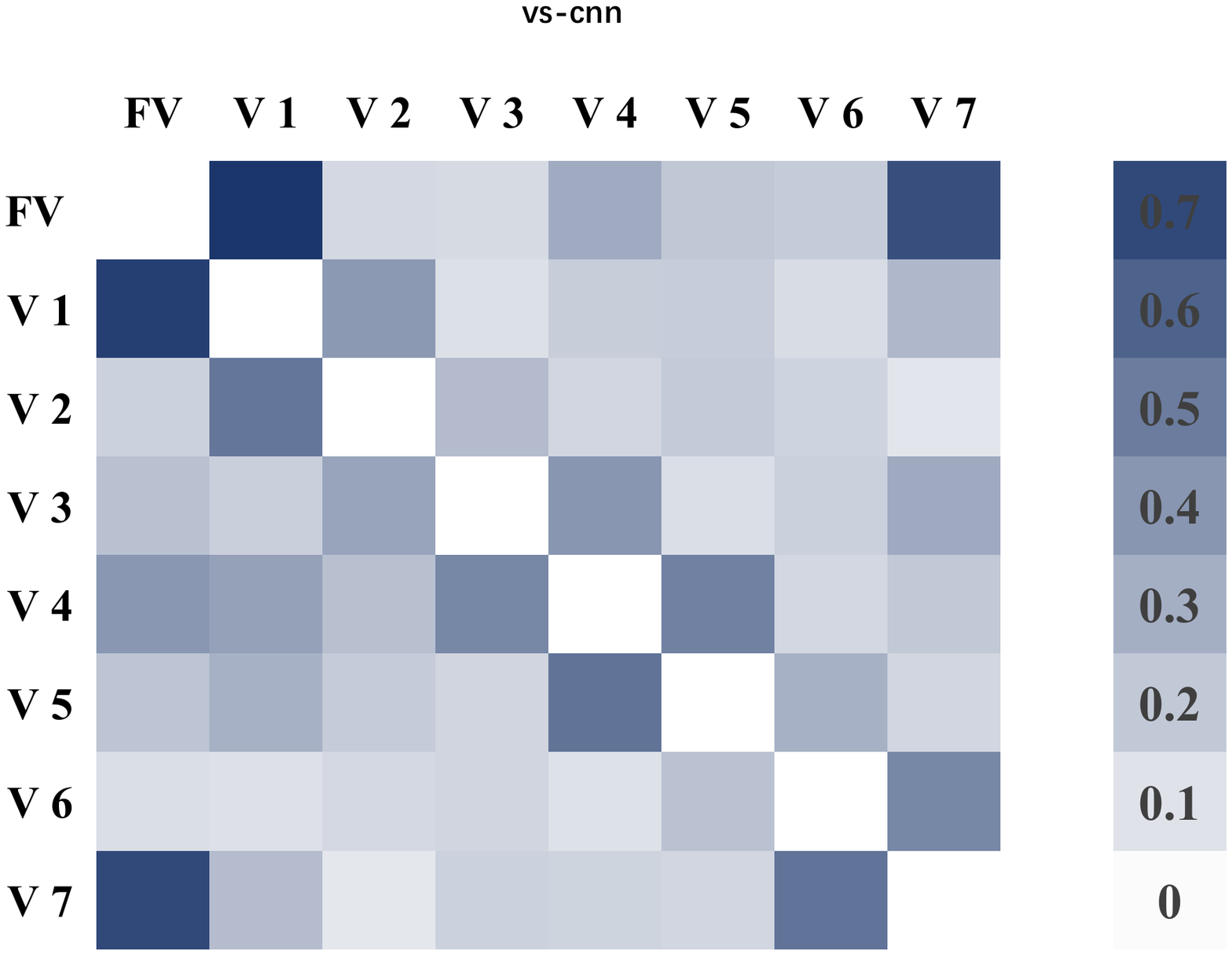}}
 \caption{Confusion matrices of the cross-view I recognition. Results of all approaches at corners are better than other positions because they correspond to the front, 1 and 7 viewpoints. In confusion matrices, rows and columns corresponding to the viewpoint 2 and 6 have lower accuracies because of skeleton distortions.}
 \label{fig:crossViewI} 
\end{figure*}

In the experiment of \textbf{Cross-view I}, we train the VS-CNN network using action samples in one of 8 fixed viewpoints, and the test is performed in other 7 fixed viewpoints. To illustrate experiment performance, we calculate the average recognition accuracy per viewpoint and build a confusion matrix including recognition results of 8 viewpoints. Figure~\ref{fig:crossViewI} shows confusion matrices of cross-view recognition which are obtained by 9 recognition approaches. In each confusion matrix, the vertical and the horizontal axis refer to training viewpoints and test viewpoints, respectively. In the figure, deep colors represent high recognition accuracies, and light colors describe lower accuracy values. We also calculate the average result of all viewpoints for each recognition approach, and show average results of 9 recognition approaches in the Table~\ref{tab:resultAll}.

As shown in the Figure~\ref{fig:crossViewI}, results of all approaches at corners are better than other positions. These results correspond to the front viewpoint, the viewpoint 1 and 7. There is a $45^\circ$ angle between the viewpoint 1 and 7 and the front viewpoint. It indicates that view change within $45^\circ$ has less effect on recognition results. Comparing 9 confusion matrices, we find that rows and columns corresponding to the viewpoint 2 and 6 have lower accuracies. The reason is that only side silhouettes are captured in the two viewpoints, thus skeletons have heavy distortions. So that recognition relying on these skeletons has worse performance. Compared with other approaches in Table~\ref{tab:resultAll}, the VS-CNN obtains a lower accuracy because samples of a single viewpoint are used in both the training and the test processing. Our model can not exploit its advantage in this situation. It is obvious that all approaches perform worse in the cross-view recognition I than the cross-subject recognition and other evaluations. That is reasonable because action samples in different viewpoints have a large variance.

The evaluation of \textbf{Cross-view II} is performed in two procedures. We separate viewpoints into two groups, where one group includes the front viewpoint, viewpoints 4, 2 and 6, and the second group contains viewpoints 1, 3, 5, and 7. Action samples of the two groups are used as training samples and test samples in turn. Results of both of the two experiments are concluded in Table~\ref{tab:crossviewII}. In the table, we list average recognition accuracies for each viewpoint in two rounds of evaluation. The final average results for all approaches are recorded in the Table~\ref{tab:resultAll}.
\begin{table*}[!t] \footnotesize
\begin{center}
\caption{Results of cross-view recognition II. Results obtained in viewpoints of the front view, viewpoints 2, 4 and 6 are a little worse because skeletons in the training set are noised.}
\label{tab:crossviewII}
\begin{tabular}{|p{1cm}|p{2.3cm}|p{1cm}|p{1cm}|p{1cm}|p{1cm}|p{1cm}|p{1cm}|p{1cm}|p{1cm}|}
\hline
 & Training & \multicolumn{4}{|c|} {V1, V3, V5, V7 } & \multicolumn{4}{|c|} {FV, V4, V2, V6} \\
\hline
 Source & Test & FV & V2 & V4 & V6  & V1 & V3 & V5 & V7 \\
\hline
\multirow{4}{*}{RGB} & JOULE~\cite{JOULE2016} & 0.74  & 0.49 & 0.57 & 0.55 & 0.74 & 0.48 & 0.47 & 0.80  \\
 & ResNeXt~\cite{ResNeXtARX2018} & 0.51  & 0.40 & 0.54 & 0.39 & 0.52  & 0.44  & 0.48  & 0.52  \\
 & C3D~\cite{C3DICCV2015} &  0.51 & 0.38  & 0.15  & 0.11  & 0.59  & 0.32  & 0.51  & 0.46   \\
 & LRCN(Resnet34)~\cite{LRCNCVPR2015} & 0.11  & 0.11  & 0.24  & 0.10 & 0.19  & 0.23  & 0.19  & 0.17  \\
 & LRCN(Resnet50)~\cite{LRCNCVPR2015} & 0.11  & 0.18  & 0.26  & 0.14  & 0.12  &  0.10 & 0.14  & 0.09  \\
 \hline
\multirow{1}{*}{Depth} & C3D~\cite{C3DICCV2015} & 0.27  & 0.18  & 0.28  &  0.10 & 0.25  & 0.19  & 0.22  & 0.23 \\
\hline
\multirow{7}{*}{Skeleton} & TCN~\cite{TCN2017} & 0.59 & 0.20 & 0.31 & 0.20 & 0.72  & 0.33  & 0.40 & 0.68 \\
 & Res-TCN~\cite{ResTCN2017} & 0.58 & 0.27 & 0.39 & 0.24 & 0.75  & 0.40  & 0.45  & 0.73  \\
 & LSTM~\cite{ShahroudyNTU2016} & 0.42 & 0.17 & 0.32 & 0.16 & 0.41 & 0.28 & 0.29 & 0.41  \\
 & P-LSTM~\cite{ShahroudyNTU2016} & 0.43 & 0.13 & 0.31 & 0.14 & 0.50  & 0.28  & 0.32  & 0.53   \\
 & SK-CNN~\cite{EnhancedSK2017} & 0.75 & \textbf{0.62} & 0.64 & 0.57 & 0.79  & \textbf{0.62}  & \textbf{0.63}  & 0.82  \\
 & ST-GCN~\cite{STGCN2018} & 0.72 & 0.32 & 0.60 & 0.29 & 0.74 & 0.52  & 0.56  & 0.74  \\
 & VS-CNN\textbf{(Ours)} & \textbf{0.87}  & 0.54  & \textbf{0.71}  & \textbf{0.60} & \textbf{0.87} & 0.58 & 0.60 & \textbf{0.87} \\
\hline
\end{tabular}
\end{center}
\end{table*}

The Table~\ref{tab:crossviewII} illustrates that results obtained in viewpoints of the front view, viewpoints 2, 4 and 6 are a little worse than the other 4 viewpoints because of noised skeletons in viewpoints of 1, 3, 5 and 7 caused by occlusions. Moreover, the results of the front view, viewpoints of 1 and 7 are better than other viewpoints as always. And nearly all approaches get worse performance in the viewpoint of 2, 3 and 6, corresponding to the view angles $90^\circ$, $135^\circ$, $225^\circ$.

According to the Table~\ref{tab:resultAll}, the VS-CNN performs much better than other approaches. The result of Cross-view II is only 5\% lower than the cross-subject recognition, which illustrates that our evaluation rule using samples of 8 fixed viewpoints to train classifiers for the arbitrary-view recognition is reasonable. Compared with the SK-CNN, though the VS-CNN has lower performance in viewpoints of 2, 3, 5 in the Table ~\ref{tab:crossviewII}, the proposed VS-CNN has a better performance than the SK-CNN considering the average result of all viewpoints, as shown in Table~\ref{tab:resultAll}. In addition, the recognition using RGB videos obtains better performance than using depth sequences in two kinds of cross-view evaluations because the depth data has only one channel so that view variance causes heavy occlusion. Referring to three action modalities,~\ie RGB videos, depth sequences, and skeleton sequences, the depth modality has the worst performance in the cross-view evaluation. It is reasonable that depth images have lower resolution than RGB frames, and it is easy to suffer heavy occlusion when the capture view changes.
\subsection{Arbitrary-view recognition}
In the evaluation of \textbf{Arbitrary-view I}, we train recognition models using action samples of 8 fixed viewpoints, and perform the test on short sections of varying-view sequences. For the \textbf{Arbitrary-view II} evaluation, we divide short sections in half according to subjects.  The half part is a training set while the other one is a testing set. They are used for model training and testing.

\begin{figure}[t]
\begin{center}
\includegraphics[width=0.8\linewidth]{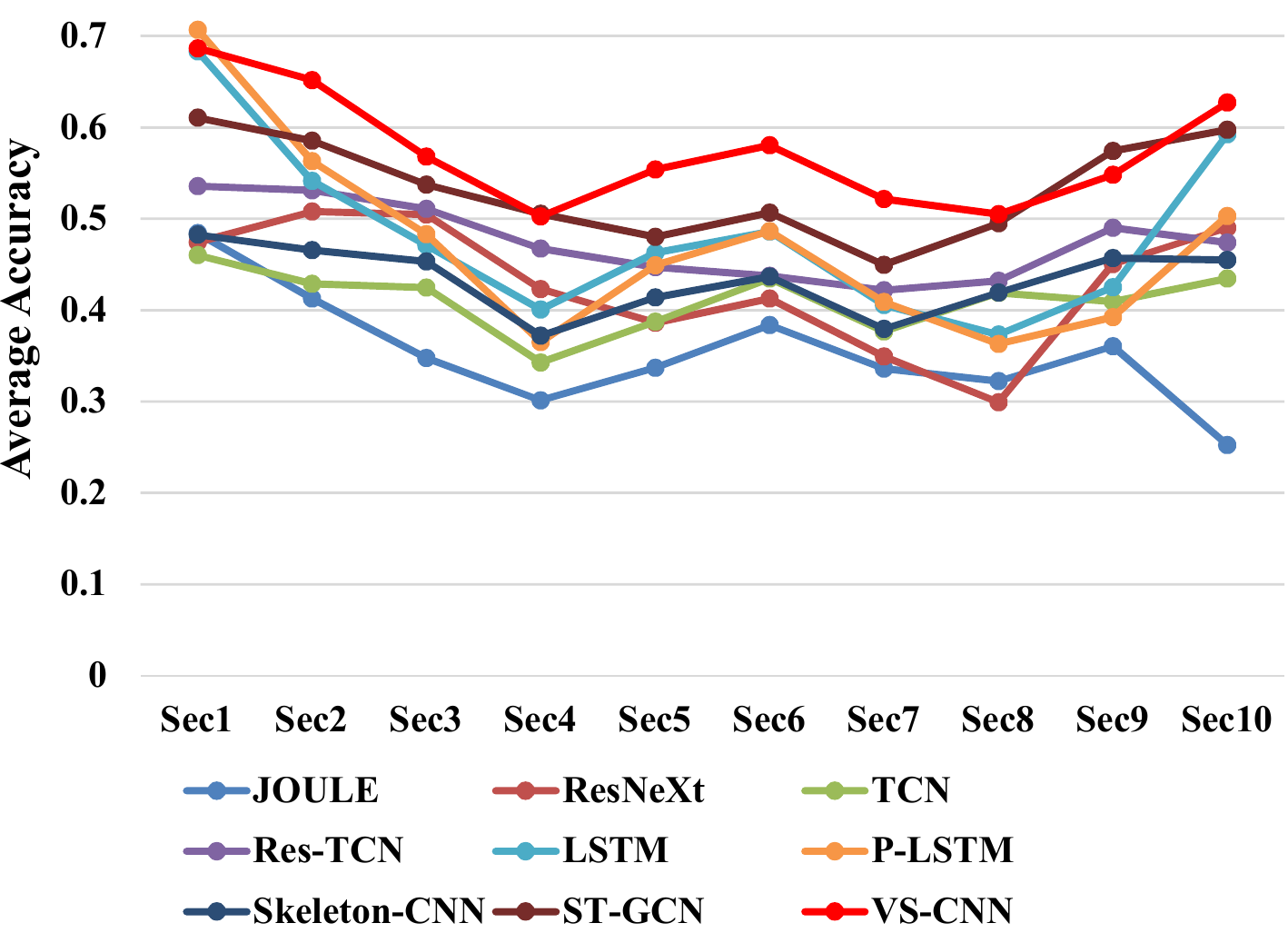}
\end{center}
   \caption{Evaluation results of the arbitrary-view recognition I. Recognition accuracies vary in a shape of ``W'' in varying-view sequences. Two valley points lie at the 4th and 7th, 8th sections, corresponding to the view range of $108^\circ \sim  144^\circ$ and $216^\circ \sim 288^\circ$ in the full-circle view. It is because occlusions lead to heavy skeleton noises.}
\label{fig:arbitrary}
\end{figure}

In the experiment of \textbf{Arbitrary-view I}, we evaluate the performance of all approaches in each temporal segmented section and show the recognition accuracy per section in Figure~\ref{fig:arbitrary}. Here, varying-view sequences are segmented to 10 clips. Since each varying-view action sequence covers the entire $360^\circ$ view angle, one separated short section covers a view angle of $36^\circ$.
The figure shows that recognition accuracies vary in a shape of ``W'' in varying-view sequences. At the beginning and the end of varying-view sequences, recognition accuracies are high for all approaches because the view angle of the moving sensor is near the front viewpoint so that captured data quality of actions is better, no distortion and no occlusion. In varying-view sequences, the recognition accuracy gradually decreases and reaches the first valley point, then it increases until a peak point, and decreases again to arrive at the second valley point. Following that, the recognition accuracy increases once again until the end of the sequence. According to the figure, the two valley points lie at the 4th and 7th, 8th sections. If the position of the front view is defined as $0^\circ$, the 4th segmented section occupies the view range of $108^\circ \sim  144^\circ$. And the 7th, 8th sections cover an angle range of $216^\circ \sim 288^\circ$. These two positions correspond to neighbor areas of the viewpoint V3 and V5 in the Figure~\ref{fig:captureSet}. In these view ranges, occlusions lead to heavy noises in skeleton sequences, and information loss in RGB videos. Therefore, recognition results are worse.

\begin{figure}[t]
\begin{center}
\includegraphics[width=0.8\linewidth]{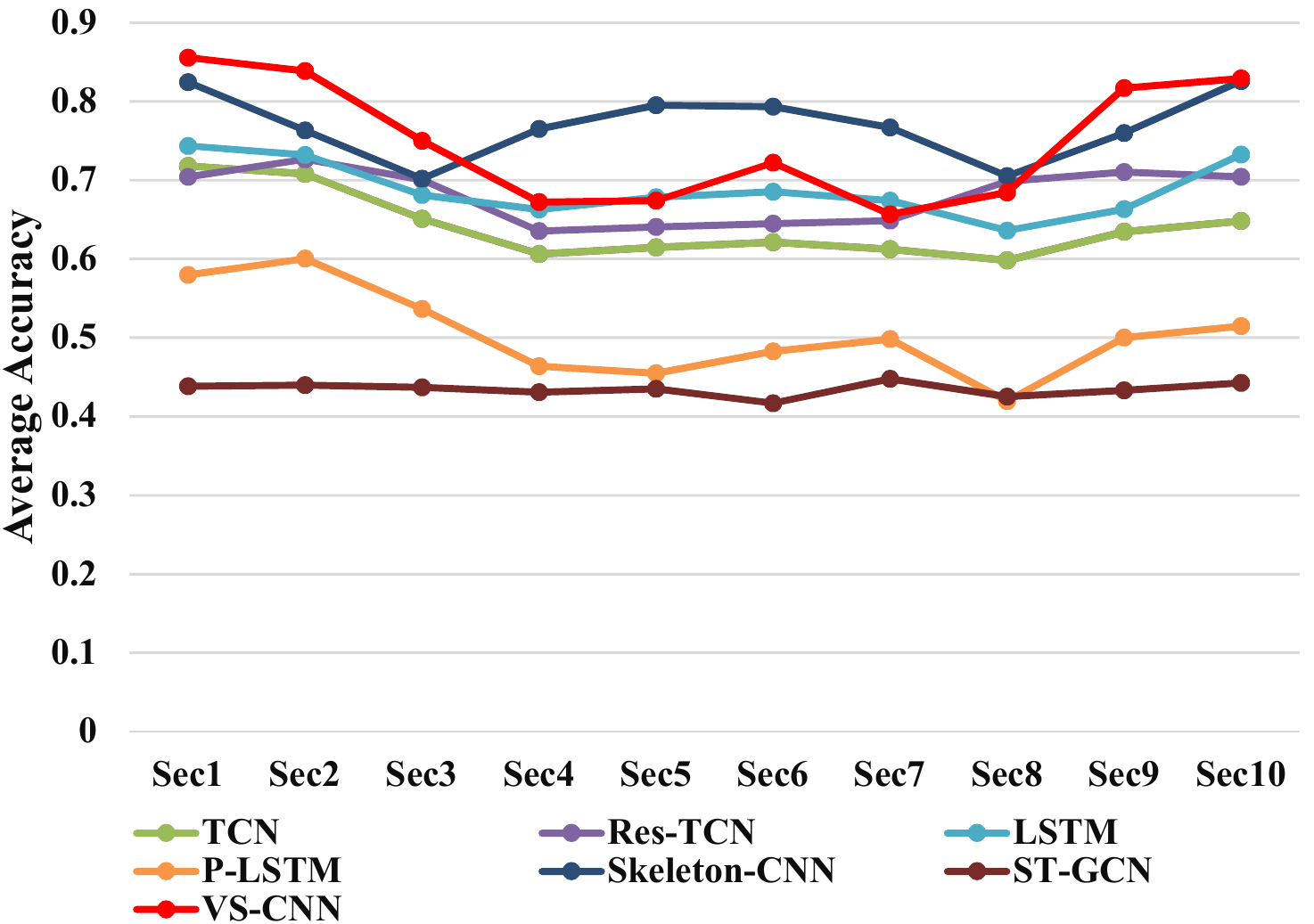}
\end{center}
   \caption{Results of the arbitrary-view recognition II. Curves of recognition accuracies have flat shapes for most of the recognition approaches. It is because both of the training set and the test set cover the full-circle views of actions.}
\label{fig:arbitraryII}
\end{figure}

For the experiment of \textbf{Arbitrary-view II}, we use action sections in the full-circle view sequences to train recognition models, and evaluate the performance of trained models on segmented varying-view sections. Figure~\ref{fig:arbitraryII} shows average recognition accuracies of 10 sections in varying-view sequences which are obtained using various recognition approaches. Since the JOULE and the ResNeXt performs worse in most above experiments, we do not list results of them here. As shown in the Figure~\ref{fig:arbitraryII}, we obtain better recognition performance using varying-view sequences to train recognition models compared with results of the arbitrary-view recognition I. Furthermore, we can see that curves of recognition accuracies have flat shapes for most of the recognition approaches in the Figure~\ref{fig:arbitraryII}, that is much different from the Figure~\ref{fig:arbitrary}. It is because both of the training set and the test set cover the full-circle views of actions which improve recognition performance.

\begin{figure}[t]
\begin{center}
\includegraphics[width=0.7\linewidth]{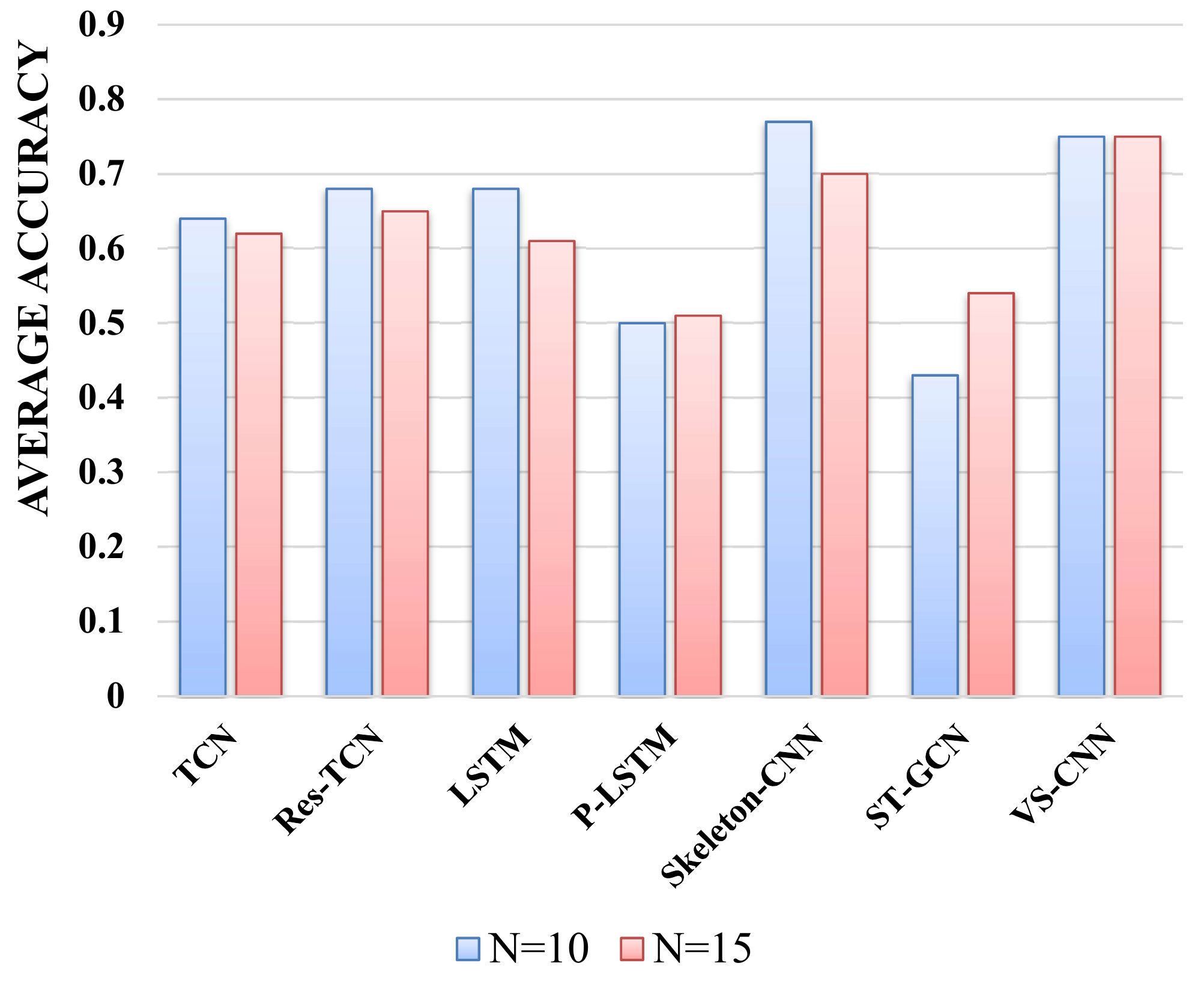}
\end{center}
   \caption{Experiment comparison between separating varying-view sequences into 10 and 15 sections in the arbitrary-view recognition II. Results show that it is a better choice to clip varying-view sequences to 10 short sections.}
\label{fig:arbitraryII1015}
\end{figure}

In addition, we further evaluate the performance of section segmentation in the experiment of Arbitrary-view recognition II. We change the number of segmented sections to 15 and evaluate the performance of all recognition approaches again in the experiment of \textbf{Arbitrary-view II}. We average recognition accuracies of all sections in varying-view sequences for all recognition approaches and compare these results with results obtained by segmenting one sequence into 10 sections in the Figure~\ref{fig:arbitraryII1015}. The result comparison shows us that it is a better choice to separate varying-view sequences into 10 sections. In this situation, segmented clips have a similar length with action sequences captured in fixed viewpoints. It is suitable for our experiments.

\subsection{Summary}
Above evaluations certified that our proposed VS-CNN network outperforms existing approaches in experiments of cross-subject recognition, cross-view recognition, and arbitrary-view recognition.
Comparing different types of evaluations, the cross-subject recognition obtains the highest recognition accuracy, and the cross-view recognition I and the arbitrary-view recognition I perform a little worse. In the cross-subject recognition, action samples have the same viewpoints in the training and the test steps, but it is a totally different situation in other three types of evaluations, especially in the cross-view I and the arbitrary-view recognition. It is mainly due to unequal data distribution in the training and the test set in these experiments. The experiment comparison between the arbitrary-view recognition I and the arbitrary-view recognition II also indicates this problem clearly. However, it is impossible to collect actions in arbitrary views in our real-world HRI applications. It is required to propose approaches to recognize arbitrary-view actions based on training samples captured in limited views.
It is still a challenging problem of action recognition with unknown viewpoints, and we will continue with the topic in our future research.

\section{Conclusions}
\label{sec:conclu}
In this paper, we newly collected a large-scale RGB-D action dataset for arbitrary-view action analysis. The dataset contains samples captured in 8 fixed viewpoints and varying-view sequences that cover the entire $360^\circ$ view angels. Samples captured in fixed viewpoints provide training data for the arbitrary-view recognition, and also may be used for the multi-view recognition. The dataset contained more viewpoints, more subjects, and especially varying-view sequences covering a full-circle $360^\circ$ view angles.
The dataset provided sufficient data for multi-view and arbitrary-view action analysis. We further proposed a VS-CNN network to recognize arbitrary-view actions, and we evaluate the proposed network for the cross-subject recognition, the cross-view recognition, and the arbitrary-view recognition on our dataset. All experiments are compared with related 8 action recognition approaches. These experiments certified the superior performance of the proposed VS-CNN network.

\section*{Acknowledgment}
This research is supported by the Natural Science Foundation of China (NSFC) under grant No. 61673088 and grant No. 61305043.
This work was partly supported by the 111 Project No. B17008.
\ifCLASSOPTIONcaptionsoff
  \newpage
\fi


\bibliographystyle{IEEEtran}
\bibliography{sample-bibliography}

\end{document}